\documentclass[10pt,twocolumn,letterpaper]{article}

\usepackage{cvpr}              

\usepackage[dvipsnames]{xcolor}

\definecolor{cvprblue}{rgb}{0.21,0.49,0.74}
\usepackage[pagebackref,breaklinks,colorlinks,citecolor=cvprblue]{hyperref}

\usepackage{wrapfig}
\usepackage{multirow}
\usepackage{graphicx}
\usepackage{pifont}
\usepackage{subcaption}

\NewDocumentEnvironment{varsubfigure}{O{c}mo}
{
	\begin{subfigure}[#1]{#2}
		\if#1t\vspace{0pt}\fi
		\makebox[0pt][r]{%
			\refstepcounter{subfigure}%
			\IfValueT{#3}{\label{#3}}%
			(\thesubfigure) 
		}%
		\begin{tabular}{@{}p{\textwidth}@{}}
		}
		{\end{tabular}\if#1b\vspace{0pt}\fi\end{subfigure}}
\def\paperID{5965} 
\def\confName{CVPR}
\def\confYear{2024}

\title{GenDet: Towards Good Generalizations for AI-Generated Image Detection}

\author{%
	Mingjian Zhu, Hanting Chen, Mouxiao Huang, Wei Li, Hailin Hu, Jie Hu, Yunhe Wang\\
	Huawei Noah’s Ark Lab\\
	\texttt{\{zhumingjian, yunhe.wang\}@huawei.com} \\}

\begin{document}
\maketitle
\begin{abstract}
The misuse of AI imagery can have harmful societal effects, prompting the creation of detectors to combat issues like the spread of fake news. Existing methods can effectively detect images generated by seen generators, but it is challenging to detect those generated by unseen generators. They do not concentrate on amplifying the output discrepancy when detectors process real versus fake images. This results in a close output distribution of real and fake samples, increasing classification difficulty in detecting unseen generators. This paper addresses the unseen-generator detection problem by considering this task from the perspective of anomaly detection and proposes an adversarial teacher-student discrepancy-aware framework. Our method encourages smaller output discrepancies between the student and the teacher models for real images while aiming for larger discrepancies for fake images. We employ adversarial learning to train a feature augmenter, which promotes smaller discrepancies between teacher and student networks when the inputs are fake images. Our method has achieved state-of-the-art on public benchmarks, and the visualization results show that a large output discrepancy is maintained when faced with various types of generators.
\end{abstract}    
\section{Introduction}
\label{sec:intro}
In the past decade, the landscape of generative models has undergone significant evolution, with a proliferation of visual generative architectures rooted in GANs (Generative Adversarial Networks)~\cite{zhu2017unpaired,park2019semantic} and diffusion models~\cite{gu2022vector,rombach2022high}. 
Early generative architectures are primarily based on GANs, which mainly utilize adversarial learning between a generator and a discriminator to enable the generator to create high-quality images, as seen in BigGAN~\cite{brock2018large} and ProGAN~\cite{karras2017progressive}. Recently, diffusion models have significantly improved the quality of generated images, as exemplified by ADM~\cite{dhariwal2021diffusion} and Stable Diffusion~\cite{rombach2021highresolution}. The working principle of diffusion models involves adding noise to available training data, followed by reversing the process to restore the data. The models progressively master noise elimination, enabling the creation of high-quality images from random noise through a refined denoising process. The fidelity of the generated images and videos has made such progress that human perceptual discernment is often challenged in distinguishing their veracity.

The advancement, while monumental, raises concerns about negative social impacts. For instance, recent misapplications of these models have been evidenced in fake news of a 9.1 magnitude earthquake and tsunami on North America's West Coast~\cite{Earthquake}. The associated image is synthesized via the AI platform, Midjourney~\cite{midjourney}. Consequently, the capacity to recognize outputs from image forgery methods becomes paramount in the prudent oversight of AI methodologies. As the human eye struggles to distinguish AI-generated images, there is an increasing inclination toward training AI detectors to recognize generated images.

\begin{figure}[t]
	\centering
	\captionsetup[subfigure]{labelformat=empty}

	\begin{subfigure}[b]{0.14\textwidth}
		\centering
		\includegraphics[width=\textwidth]{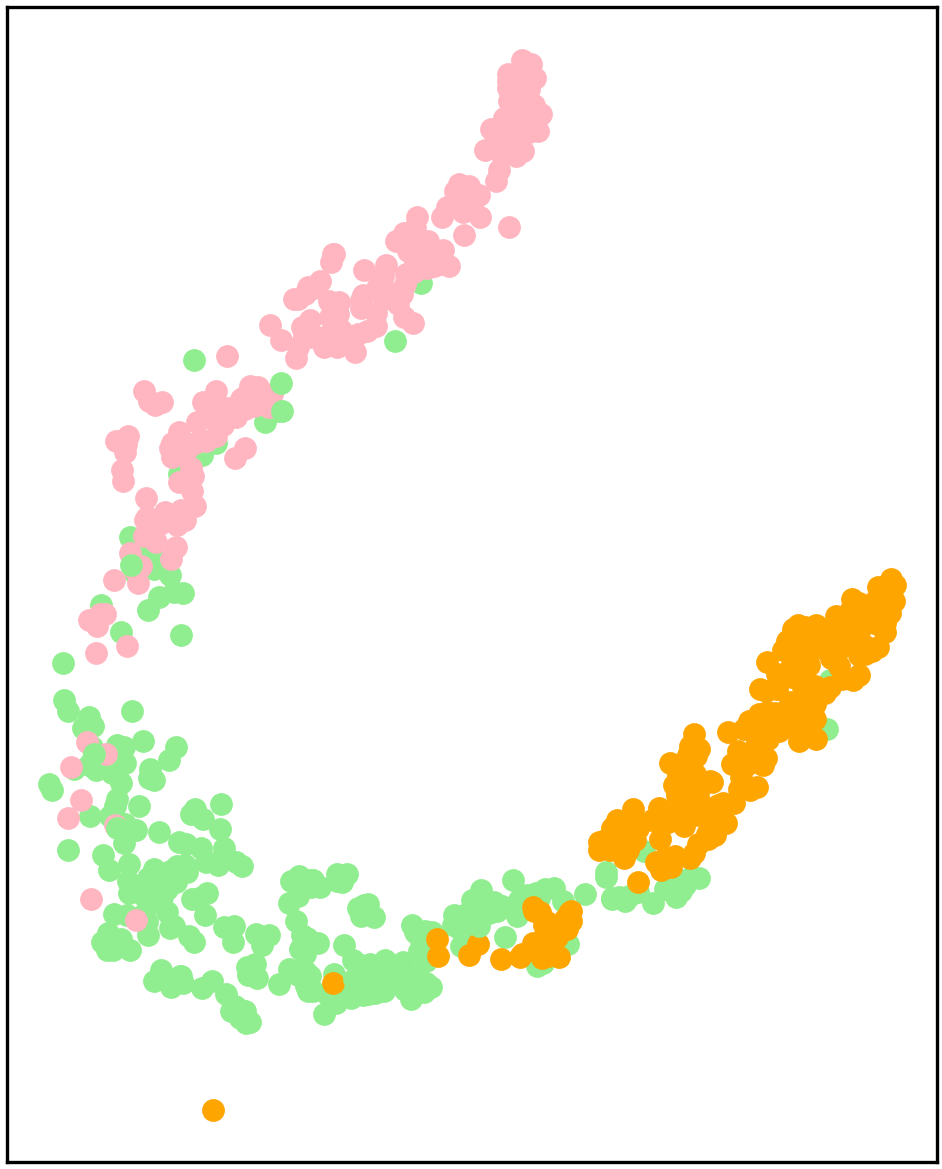}
		\caption{CNNDet~\cite{wang2020cnn}}
	\end{subfigure}
	\hspace{2mm}
	\begin{subfigure}[b]{0.14\textwidth}
		\centering
		\includegraphics[width=\textwidth]{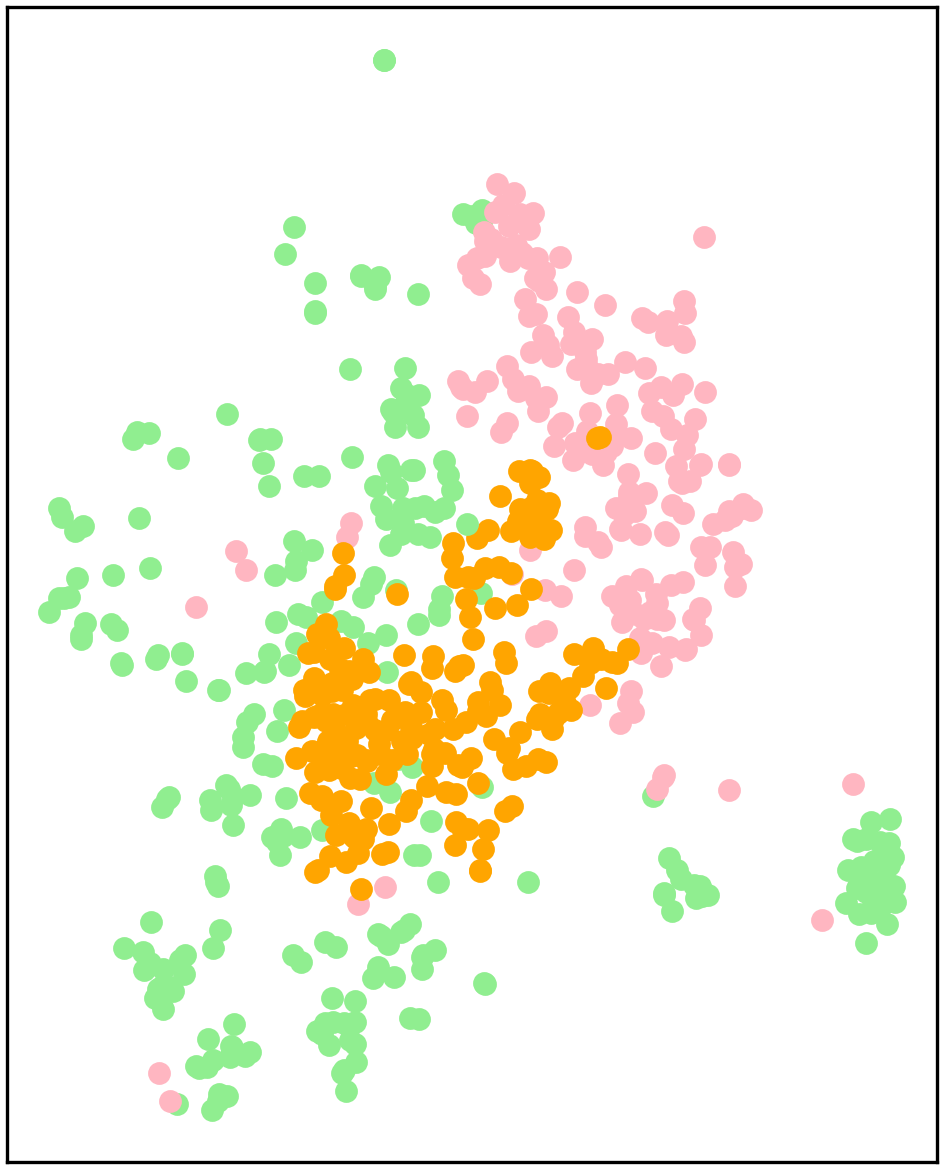}
		\caption{Ojha~\etal~\cite{ojha2023towards}}
	\end{subfigure}
	\hspace{2mm}
	\begin{subfigure}[b]{0.14\textwidth}
	\centering
	\includegraphics[width=\textwidth]{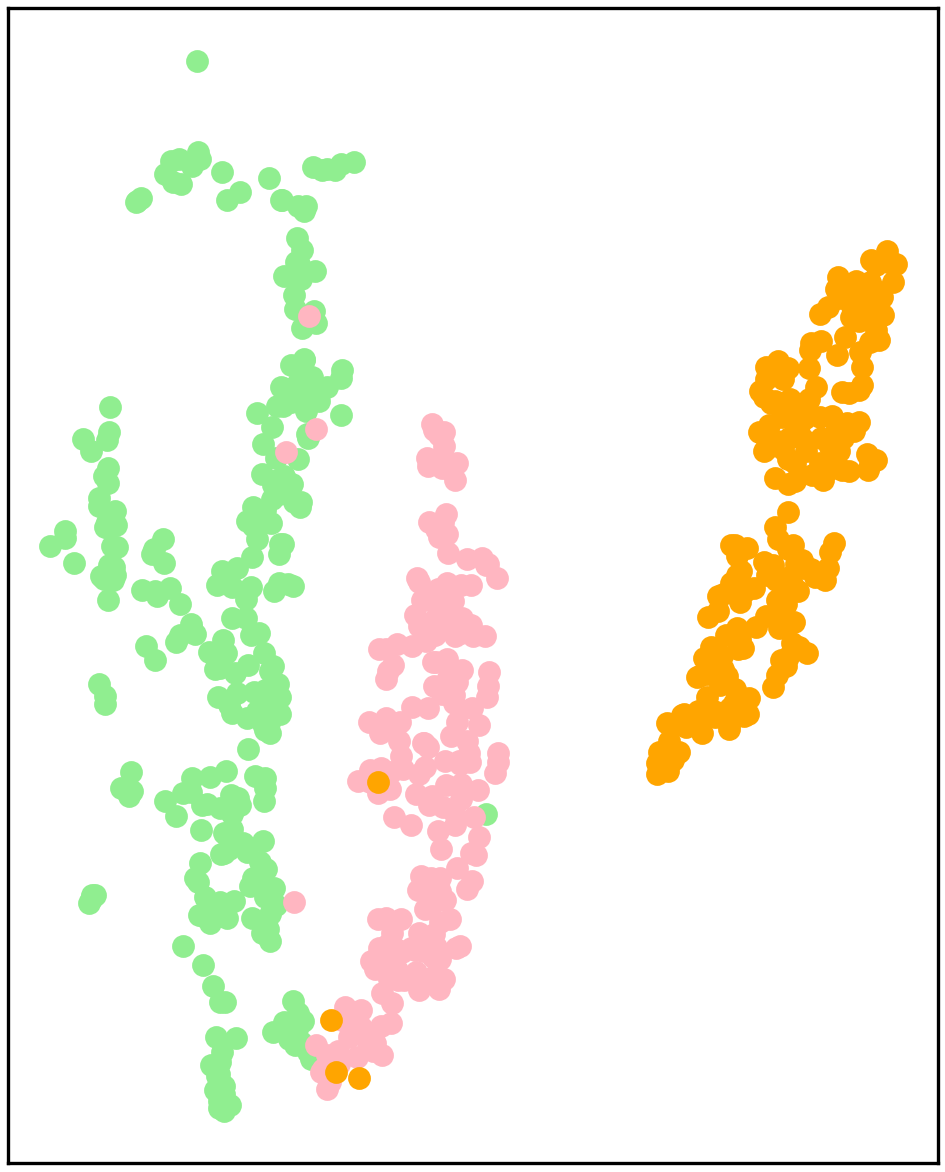}
	\caption{GenDet (Ours)}
\end{subfigure}
	\caption{We use t-SNE to visualize real and fake images on the UniversalFakeDetect dataset~\cite{ojha2023towards}. The real images are shown in red. The fake images generated by seen and unseen generators are shown in orange and green, respectively. A closer distribution of real and fake images increases the difficulty of classification.}
\label{tsne:methods}
\end{figure}



A straightforward method to identify real and fake images is using cross-entropy loss to train a neural network for binary classification~\cite{ojha2023towards, hemorrhage, xi2023ai}. However, this approach has been found to achieve high performance on seen generators and low performance on unseen generators~\cite{wang2020cnn, zhu2023genimage}. A series of methods are proposed to alleviate this problem. CNNDet~\cite{wang2020cnn} utilizes an image augmentation method to improve the recognition performance of the binary classifier. They apply hand-crafted image augmentation methods such as Gaussian Blur and JPEG compression on training data, which improves the generalization and robustness capabilities. Ojha~\etal~\cite{ojha2023towards} takes advantage of the fixed feature space of the pre-trained CLIP model~\cite{radford2021learning} to improve its generalization performance to unseen generators. These methods do not put emphasis on enlarging the output discrepancy when inputting the real and fake images into the detector. This leads to a close output distribution of real and fake images, which increases the difficulty of classification, especially when facing unknown generators.

We consider AI-generated image detection from a different perspective, namely, approaching the task with the methodology of anomaly detection. The fake images can originate from unknown distributions, and our objective is to identify all fake images that differ from true images, essentially seeking out anomalies. Among the anomaly detection methods, the teacher-student based approach~\cite{salehi2021multiresolution, bergmann2020uninformed, deng2022anomaly} has been readily validated as an effective solution for anomaly detection tasks in terms of distinguishing between normal data and abnormal data. In this approach, only the normal samples are utilized to minimize the output discrepancy between teacher and student, and the abnormal data is unseen during training. Thus, the output discrepancy will be small when fed with normal samples and most likely large when fed with various anomalous samples in inference. Intuitively, the teacher-student anomaly detection framework is a good fit for our task. However, directly applying this framework to AI-generated image detection has not fully exploited the message from fake images in training.


In this paper, we propose an adversarial teacher-student discrepancy-aware framework named GenDet for AI-generated image detection. With the goal of developing a detector that can effectively generalize to unseen image generators, we propose to jointly perform teacher-student discrepancy-aware learning and generalized feature augmentation. In particular, we first train a binary classifier as the teacher. Based on teacher-student learning schemes, we input real images and train the student to minimize the output discrepancy with the teacher. We also utilize the fake images in the training set. We maximize the output discrepancy between the student and teacher with augmented fake images as input. Generalized feature augmentation aims to generalize features extracted by fake images from seen generators. The training objective of the feature augmenter is to minimize the discrepancy between the outputs derived from the teacher and the student. This increases the probability of observing a large discrepancy when unseen fake images are fed in inference. The output of the detector is based on the discrepancy between the output of the teacher and the student. By iteratively training the above three stages in an adversarial learning manner, the resulting output of the detector would be small with real images as input and large with images generated by seen or unseen generators as input. The increased differences in the output of the detector enable easier detection. As shown in Figure~\ref{tsne:methods}, we visualize the output distribution of different detectors. The outputs of the existing methods are hard to separate when those of our detector are easily separated. Our method has achieved state-of-the-art results on public benchmarks. For example, on the GenImage~\cite{zhu2023genimage} and UniversalFakeDetect Dataset~\cite{ojha2023towards}, we have surpassed the existing state-of-the-art method by 6.8\% and 13.0\% in average accuracy, respectively.

\section{Related Work}
\label{sec:related_work}

\subsection{AI-Generated Image Detection}

The existing AI-generated images have reached a level of realism that makes them indistinguishable from the human eye. This advanced capability presents a potential risk, as the technology could be exploited to produce fake news. To avoid the risk, a series of datasets~\cite{bird2023cifake, verdoliva2022, zhu2023genimage} and approaches~\cite{chai2020makes, zhang2019detecting, nataraj2019detecting} have been proposed. Ojha~\etal~\cite{ojha2023towards} uses the feature space of a large pretrained vision-language model and demonstrates an improved generalization performance on unseen generators. Patchfor~\cite{chai2020makes} proposes to use classifiers with limited receptive fields to put emphasis on local artifacts instead of the global semantics of the input images. These methods do not focus on enlarging the output discrepancy of the detector when faced with real and fake images, which makes it hard to identify fake images, especially those generated by unseen generators. We propose a method to enlarge the discrepancy. When faced with real images, the outputs of teacher and student are similar. The outputs of teachers and students become different when faced with fake images.


\begin{figure*}[ht!]
	\centering
	\begin{varsubfigure}{0.2\textheight}[fig2:a]
		\includegraphics[height=0.2\textheight]{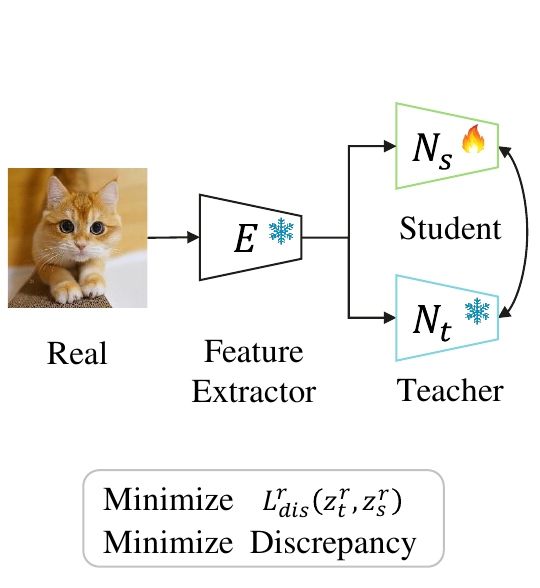}
	\end{varsubfigure}
	\quad\quad\quad
	\begin{varsubfigure}{0.2\textheight}[fig2:b]
		\includegraphics[height=0.2\textheight]{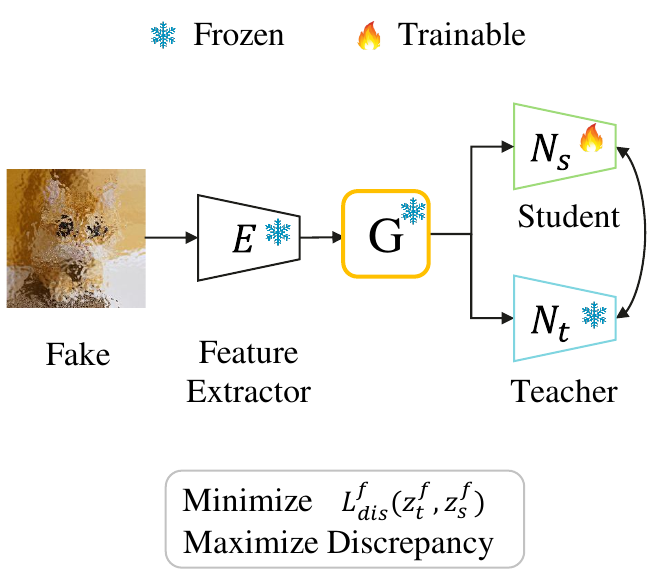}
	\end{varsubfigure}
	\quad\quad\quad\quad
	\begin{varsubfigure}{0.2\textheight}[fig2:c]
		\includegraphics[height=0.2\textheight]{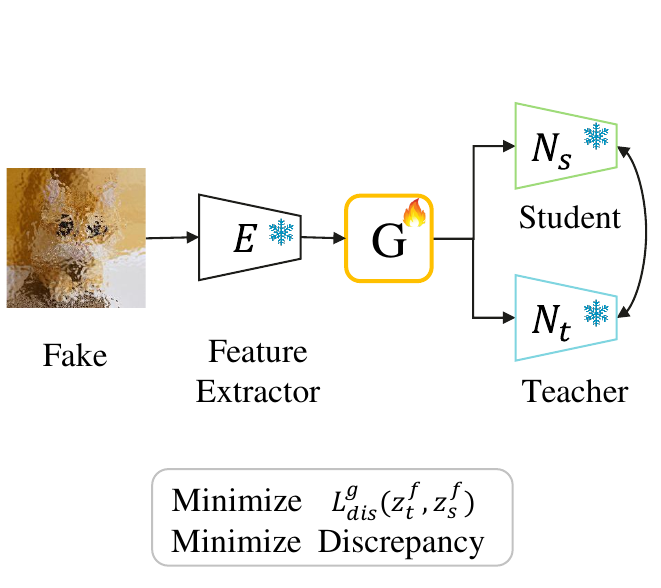}
	\end{varsubfigure}	
	\caption{Adversarial teacher-student discrepancy-aware framework. These three stages are trained in rotation. The feature extractor and the teacher network are fixed after pretraining. The student network and feature augmenter are trained in an adversarial manner.}
	\label{fig2}
\end{figure*}

\subsection{Teacher-Student Anomaly Detection}

Teacher-student based anomaly detection methods~\cite{salehi2021multiresolution,deng2022anomaly,bergmann2020uninformed} have demonstrated their effectiveness on several anomaly detection datasets~\cite{bergmann2019mvtec, braintumor, hemorrhage}. Teachers and students as a whole can be considered as a detection network. This anomaly detection method works by exploiting the differences in the output of the detection network when encountering normal and abnormal samples. During the training stage, normal samples are inputted into the detection network, and the output of the student mimics that of the teacher. The student has not seen abnormal samples in the training stage. This results in a significant discrepancy between the outputs of students and teachers when encountering abnormal samples, while a smaller discrepancy is observed with normal samples in the inference stage. Utilizing the discrepancy between the outputs of the teacher and student as the output of the detection network, the distinctiveness of the discrepancy enables the network to effectively identify anomalies. Based on this idea, MKD~\cite{salehi2021multiresolution} proposes to distill features at various layers of the teacher network into a simpler student network. This method enjoys the benefits of a more comprehensive knowledge transfer. RD~\cite{deng2022anomaly} addresses the non-distinguishing filter problem in the previous knowledge distillation models and proposes a reverse knowledge distillation strategy. This anomaly detection method has been applied to other tasks with beneficial results. OCKD~\cite{li2022one} utilizes a teacher-student framework to enhance the cross-domain performance of face presentation attack detection with one-class domain adaptation. Considering the successful practices on these tasks, we take advantage of the strengths of teacher-student anomaly detection methods and improve them for application to AI-generated image detection tasks.

\section{Method}
\label{sec:method}

\subsection{Problem Definition and Method Overview}
In this section, we introduce our proposed method, including teacher-student discrepancy-aware learning and generalized feature augmentation. We first define the problem in this paper. The dataset contains real images and fake images. The images $I$ and its label $Y$ can be organized as pairs $P_{i} = \{I_{i}, Y_{i}\}$, where $i$ denotes the index of the sample. Our objective is to develop a model with robust generalization capabilities, ensuring it delivers strong binary classification performance on test data. The model needs to determine if the test image is real or fake, regardless of whether the test image is generated by seen or unseen generators in the training set. 

We propose an adversarial teacher-student discrepancy-aware framework as shown in Figure~\ref{fig2}, which is an adversarial training scheme alternating between teacher-student discrepancy-aware learning and generalized feature augmentation. The teacher-student discrepancy-aware learning minimizes the output discrepancy between the teacher network $N_{t}$ and student network $N_{s}$ when real images are taken as input and maximizes the discrepancy when fake images are taken as input. The generalized feature augmentation uses a generalized feature augmenter $G$ to generate generalized features, which aims to minimize the discrepancy when fake images are taken as input. After training the teacher and the student network, we train a binary classifier using discrepancy between the teacher and student. In inference, the generalized feature augmenter is abandoned. The image is inputted to the feature extractor, and then the feature is inputted to the teacher and student network to compute discrepancy. The binary classifier inputs the discrepancy and gives the prediction.

\subsection{Teacher-Student Discrepancy-Aware Learning}
In contrast to directly fitting a binary classification model to a generator for training, we propose teacher-student discrepancy-aware learning, which aims to better constrain the discrepancy of teachers and students and make the real and fake images generated by various types of generators easier to distinguish. For each image, we first use a pretrained feature extractor $E$ to obtain the image feature $x=E(I)$ and use the feature for subsequent model training. We denote the feature of the real image as $x^r=E(I^r)$ and that of the fake image as $x^f=E(I^f)$, where $I^r$ denotes the real image and $I^f$ denotes the fake image. We first train a teacher network $N_{t}$ for binary classification on the training set using the cross-entropy loss function. The teacher network can classify the image as real or fake. Then, $N_{t}$ is fixed, and we train the student network $N_{s}$ using two discrepancy losses with the guidance of teacher network. Note that the two losses are not trained at the same time. As shown in Figure~\ref{fig2} (a), we feed real features $x^{r}$ into the teacher network and derive the output $z_{t}^{r}=N_{t}(x^{r})$. We also input the features $x^{r}$ to the student network and obtain the output $z^{r}_{s}=N_{s}(x^{r})$. In order to make real and fake images easier to distinguish, we encourage a small discrepancy between the output of $N_{t}$ and $N_{s}$ when real images are taken as the input. The small discrepancy is achieved by applying a target function $L_{dis}^{r}$:
\begin{align}
	\begin{aligned}
		L_{dis}^{r}(z^{r}_{t}, z^{r}_{s}) &= \frac{1}{B} \sum_{b=1}^{B}( z^{r}_{t}- z^{r}_{s})^{2}\\
		&=\frac{1}{B} \sum_{b=1}^{B}( N_{t}(x^{r})- N_{s}(x^{r}))^{2}
	\end{aligned}
\label{eq:loss_real}
\end{align}
where $B$ is the batch size. $b$ denotes the image index of a batch. We fix the teacher network and train the student network. After optimizing the student network for the input real samples, the discrepancy of $N_{t}$ and $N_{s}$ is small when fed with real image. To easily distinguish real and fake image, the discrepancy of $N_{t}$ and $N_{s}$ should be large when the input is a fake image. We use the following loss to ensure this, as shown in Figure~\ref{fig2} (b). This produces a significant difference between the discrepancy of the two when confronted with the real image and the fake image in the inference stage. We fix the teacher and feature augmenter. In order to encourage the outputs of both networks to be similar, we minimize the loss function as follows:
\begin{align}
	\small
	\begin{aligned}
		L_{dis}^{f}(z^{f}_{t},& z^{f}_{s}) = [M - |  \frac{ z^{f}_{t}} {|| z^{f}_{t}||_{2}} - \frac{   z^{f}_{s} }{|| z^{f}_{s}||_{2}} |^{2}, 0]_{+} \\
		&= [M - |  \frac{N_{t}(G(x^{f}))} {||N_{t}(G( x^{f}))||_{2}} - \frac{ N_{s}(G(x^{f})) }{||N_{s}(G(x^{f}))||_{2}} |^{2}, 0]_{+} \\
	\end{aligned}
\label{eq:loss_fake}
\end{align}
where$[.]_{[+]}= max(\text{-}, 0)$, $z^{f}_{t} = N_{t}(G(x^{f}))$, and $z^{f}_{s} = N_{s}(G(x^{f}))$. We take a hyperparameter margin $M$ as a regularization, which reflects the teacher and student desirable minimal gap between the augmented and existing fake data. When the discrepency is larger than $M$, the loss becomes 0. The features of the fake images undergo feature augmentation during the training stage. Through this loss, the large output discrepancy between teacher and student can most likely be retained when fake images are taken as input.

\subsection{Generalized Feature Augmentation}
The method of utilizing fake samples to improve the anomaly detection capabilities of the teacher-student model can result in small output discrepancies when confronting real images while increasing the discrepancy in outputs when faced with fake images. However, an issue remains: this significant output discrepancy may only be present when dealing with fake images generated by generators already encountered in the dataset and may not be maintained with unseen generators. This implies that the generalizability of the detector needs to be enhanced. The existing works~\cite{tran2021data} show that using adversarial learning to augment input data/embedding leads to better generalization in multiple tasks, including language understanding~\cite{wang2019improving}, and vision-language modeling~\cite{gan2020large}, and image classification~\cite{xie2020adversarial}. For adversarial learning, the network basically contains a generator and a discriminator. The generator is used to generate the image, while the discriminator is used to discriminate whether the image is a real image or a generated fake image.  

Inspired by adversarial learning, we formulate generalized feature augmentation into an adversarial learning framework, as shown in Figure~\ref{fig2} (c). We propose a feature augmenter similar to the generator. In detail, we fix teacher $N_{t}$  and student $N_{s}$ and train the generalized feature generator $G$. The discrepancy between the output of the teacher and student is the adversarial guidance. When the discrepancy is large, such feature augmentation has already been seen by existing generators in the training set. To encourage the generated features to be more different from the features generated by the existing generator in the training set, we train the feature augmenter to generate input features for teachers and students, which makes the output discrepancy as small as possible. The augmenter increases the difficulty of keeping large output discrepancy when fake images are input to the teacher and student. This helps when training students in the previous stage so that they are able to maintain a large discrepancy from the output teacher network for all kinds of fake inputs. Compared to the design of an image augmenter, where the image is first fed into the augmenter and then into the feature extractor, the use of a feature augmenter design allows for the output of the feature by the extractor to be stored before the training stage. Using the stored features enables the training of both teacher and student models without the need for forward computation by the computationally heavy feature extractor, thereby saving on the training costs of the model. To train a feature augmenter for adversarial learning, the loss $L_{dis}^{g}$ can be formulated as:
\begin{align}
	\begin{aligned}
		L_{dis}^{g}(z^{f}_{t},  z^{f}_{s}) &= \frac{1}{B} \sum_{b=1}^{B}( z^{f}_{t}-  z^{f}_{s})^{2} \\
		&=\frac{1}{B} \sum_{b=1}^{B}( N_{t}(G(x^{f}))- N_{s}( G( x^{f})))^{2}\\
	\end{aligned}
\label{eq:loss_aug}
\end{align}
Once the teacher and student are trained, we input the discrepancy of teacher and student to train a binary classifier $N_{c}$, and its output is the prediction of real and fake images. The discrepancy $D^{r}$ and $D^{f}$ of the real and fake images for training the binary classifier are shown in Equation~\ref{eq:dis}.
\begin{align}
	\begin{aligned}
D^{r} = (z_{t}^{r}- z_{s}^{r})^{2}  \quad D^{f} = (z_{t}^{f}- z_{s}^{f})^{2}
\end{aligned}
\label{eq:dis}
\end{align}

In the inference stage, the input images are not input to the feature augmenter after feature extraction but are directly fed to the teacher and student to compute their discrepancy, and the binary classifier $N_{c}$ gives a final prediction $Y_{p}$ of real and fake images:
\begin{align}
	\begin{aligned}
		Y_{p} = N_{c}((N_{t}(x) - N_{s}(x))^{2})
	\end{aligned}
	\label{eq:clssifier}
\end{align}

\section{Experiments}
\label{sec:experiments}
\subsection{Datasets and Experimental Settings}
\paragraph{Datasets.} To assess the efficacy of our proposed GenDet, we conducted experiments using the UniversalFakeDetect Dataset~\cite{ojha2023towards} and the GenImage Dataset~\cite{zhu2023genimage}. The UniversalFakeDetect Dataset is predominantly composed of GAN-generated images and is built upon the foundation of the ForenSynths Dataset~\cite{wang2020cnn}. The latter comprises 720K images, with an equal split of 360K real and 360K fake images. The training set of the UniversalFakeDetect Dataset retains its foundation from the ForenSynths Dataset, \ie using ProGAN as the generator. In the test set, the UniversalFakeDetect Dataset contains the generators in the ForenSynths Dataset: ProGAN~\cite{karras2017progressive}, CycleGAN~\cite{zhu2017unpaired}, BigGAN~\cite{brock2018large}, StyleGAN~\cite{karras2019style}, GauGAN~\cite{park2019semantic}, StarGAN~\cite{choi2018stargan}, Deepfakes~\cite{rossler2019faceforensics++}, SITD~\cite{chen2018learning}, SAN~\cite{dai2019second}, CRN~\cite{chen2017photographic}, IMLE~\cite{li2019diverse}. Besides, UniversalFakeDetect Dataset has been augmented with three kinds of diffusion models (Guided Diffusion Model~\cite{dhariwal2021diffusion}, GLIDE~\cite{nichol2021glide}, and LDM~\cite{rombach2022high}) and one autoregressive model (DALL-E~\cite{ramesh2021zero}), building upon the foundation of ForenSynths.

The GenImage dataset primarily employs the Diffusion model for image generation. Drawing from the ImageNet dataset, GenImage utilizes real images and their associated labels to produce various fake images. The training set of GenImage contains fake images generated using Stable Diffusion V1.4~\cite{rombach2022high}. For the test set, we incorporate a variety of generators, including Stable Diffusion V1.4~\cite{rombach2022high}, Stable Diffusion V1.5~\cite{rombach2022high}, GLIDE~\cite{nichol2021glide}, VQDM~\cite{gu2022vector}, Wukong~\cite{wukong}, BigGAN~\cite{brock2018large}, ADM~\cite{dhariwal2021diffusion}, and Midjourney~\cite{midjourney}. In total, GenImage comprises 1,331,167 real and 1,350,000 fake images. Following the settings on GenImage, we perform a cross-generator image classification task using all images generated by Stable Diffusion V1.4 as well as the corresponding portion of real images to train the detector. We test the detector on all kinds of generators. We also evaluate the degraded image classification task on this dataset.

\paragraph{Implementation Details.} The structure of the teacher network, student network, and feature augmentor is a transformer block. The structure of the classifier model is a transformer block and a fully-connected layer. The dimension of the transformer is 196. The transformer contains one layer and four heads. The expansion rate of feed forward network in transformer block is 4. The learning rate for training the teacher on UniversalFakeDetect and GenImage is 0.00001. The learning rates for training the student on UniversalFakeDetect and GenImage are 0.0001 and 0.00001, respectively. The learning rates for training the classifier on UniversalFakeDetect and GenImage are 0.00001 and 0.001, respectively. In the adversarial teacher-student discrepancy-aware framework, training rotates through three stages, each for one epoch sequentially. Three stages can be considered as one rotation. The number of rotations is 10. The optimizer is AdamW. We use CLIP:ViT~\cite{radford2021learning}, a feature extractor pretrained on a large dataset of 400M imaget-text pairs, to extract features from the image.

\paragraph{Evaluation Metrics.} Following previous approaches for AI-generated image detection~\cite{wang2020cnn,ojha2023towards,wang2023dire}, we evaluate the performance of detectors using mean average precision (mAP) and classification (average accuracy) as evaluation metrics. For the UniversalFakeDetect dataset, mAP and average accuracy are used to evaluate the detectors. We also follow UniversalFakeDetect dataset to compute the threshold. For the GenImage dataset setting, average accuracy is used to evaluate the effect, and the threshold is set to 0.5.

\begin{table*}[t!]
	{\small
		\centering
		\tabcolsep=0.1cm
		\resizebox{1.\linewidth}{!}{
			\begin{tabular}{cc cccccc c cc cc c ccc ccc c c}
				\toprule
				
				\multirow{2}{*}{\shortstack[c]{Detection\\method}}  & \multicolumn{6}{c}{Generative Adversarial Networks} & \multirow{2}{*}{\shortstack[c]{Deep\\fakes}} & \multicolumn{2}{c}{Low level vision} & \multicolumn{2}{c}{Perceptual loss} &\multirow{2}{*}{Guided} & \multicolumn{3}{c}{LDM} & \multicolumn{3}{c}{Glide} & \multirow{2}{*}{DALL-E} & Total \\
				\cmidrule(lr){2-7} \cmidrule(lr){9-10} \cmidrule(lr){11-12} \cmidrule(lr){14-16} \cmidrule(lr){17-19} \cmidrule(lr){21-21}

				& \shortstack[c]{Pro-\\GAN} & \shortstack[c]{Cycle-\\GAN} & \shortstack[c]{Big-\\GAN} & \shortstack[c]{Style-\\GAN} & \shortstack[c]{Gau-\\GAN} &  \shortstack[c]{Star-\\GAN}   &  & SITD & SAN & CRN & IMLE & & \shortstack[c]{200\\steps} & \shortstack[c]{200\\w/ CFG} & \shortstack[c]{100\\steps} & \shortstack[c]{100\\27} & \shortstack[c]{50\\27} & \shortstack[c]{100\\10} & & \shortstack[c]{mAP(\%)}
				
				\\ 
				
				\midrule
				
				{\shortstack[c]{CNNDet}}~\cite{wang2020cnn}  & \textbf{100.0} & 93.47 & 84.50 & {99.54} & 89.49 & 98.15 & 89.02 & {73.75} & 59.47 & {98.24} & 98.40 & 73.72 & 70.62 & 71.00 & 70.54 & 80.65 & 84.91 & 82.07 & 70.59 & 83.58\\
				
				{\shortstack[c]{Patchfor}}~\cite{chai2020makes}
				&  80.88 & 72.84 & 71.66 & 85.75 & 65.99 & 69.25 & 76.55 & \textbf{76.19} & {76.34} & 74.52 & 68.52 & 75.03 & 87.10 & 86.72 & 86.40 & 85.37 & 83.73 & 78.38 & 75.67 & 77.73\\
				
				
				{\shortstack[c]{Co-occurence}}~\cite{nataraj2019detecting}   &  99.74 & 80.95 & 50.61 & 98.63 & 53.11 & 67.99 & 59.14& 68.98 & 60.42 & 73.06 & 87.21 & 70.20 & 91.21 & 89.02 & 92.39 & 89.32 & 88.35 & 82.79 & 80.96 & 78.11\\
				
				
				\shortstack[c]{Spec}~\cite{zhang2019detecting}  & 55.39 & \textbf{100.0} & 75.08 & 55.11 & 66.08 & \textbf{100.0} & 45.18 & 47.46 & 57.12 & 53.61 & 50.98 & 57.72 & 77.72 & 77.25 & 76.47 & 68.58 & 64.58 & 61.92 & 67.77 & 66.21\\
				
				\shortstack[c]{DIRE$^{*}$}~\cite{wang2023dire} & 100.0 &76.73 & 72.8 & 97.06& 68.44 & 100.0& 98.55& 54.51 &65.62 & 97.10 & 93.74 & 94.29& 95.17& 95.43&95.77& 96.18 &97.30 & 97.53 &68.73 &87.63\\
				
				{\shortstack[c]{{Ojha~\etal$^{*}$}}}~\cite{ojha2023towards}  & \textbf{100.0} & 99.46 & {99.59} & 97.24 & \textbf{99.98} & 99.60 & 82.45 & 61.32 & {79.02} & 96.72 & \textbf{99.00} & {87.77} & {99.14} & {92.15} & {99.17} & {94.74} & {95.34} & 94.57 & {97.15} & {93.38}\\ \hline
				{\shortstack[c]{{Teacher}}}  & \textbf{100.0} & {95.65} & {98.38} & {96.10}& {99.94} & 92.86 & {77.68} & 72.18 & \textbf{81.63} & 94.43 & 99.00 & {92.26} & {98.98} & {92.25} & {99.16} & {97.11} &{96.77} &{96.38} & 
				{93.80} & {93.40}\\		
				
				{\shortstack[c]{{Teacher-Student AD}}}  & {99.86} & {99.87} & {99.79} & {99.69}& {99.73} & 98.64 & {84.97} & 62.21 & 54.21 & 90.98 & 98.71 & {98.89} & {99.53} & {98.92} & {99.51} & {99.17} &{99.16} &{98.99} & {98.69} & {93.76}\\	
				
				{\shortstack[c]{{Teacher-Student AD w/fake}}}  & {99.92} & {99.83} & {99.70}& {99.65} & 99.69 & {97.79} & \textbf{91.71} & 56.08 & 56.68 & \textbf{98.61} & {98.87} & {98.96} & {99.32} & {99.12} & {99.33} &{98.99} &{99.03} &{98.92} & {99.01} &94.27\\	
				
				{\shortstack[c]{{GenDet}}}  & {99.95} & {99.95} & \textbf{99.92} & \textbf{99.92}& {99.92} & 99.25 & {91.38} & 61.23 & 72.66 & 97.90 & 98.88 & \textbf{99.30} & \textbf{99.85} & \textbf{99.51} & \textbf{99.85} & \textbf{99.50} &\textbf{99.46} &\textbf{99.19} & \textbf{99.47} & \textbf{95.64}\\	
				\bottomrule
		\end{tabular}}
	}
	\caption{Results on UniversalFakeDetect dataset. The methods are compared with the mAP as an evaluation metric.}
	\label{tab:Performance_Comparison_on_UniversalFakeDetect_ap}
	
\end{table*}

\begin{table*}[t!]
	{\small
		\centering
		\tabcolsep=0.1cm
		\resizebox{1.\linewidth}{!}{
			\begin{tabular}{cc cccccc c cc cc c ccc ccc c c}
				\toprule
				
				\multirow{2}{*}{\shortstack[c]{Detection\\method}}  & \multicolumn{6}{c}{Generative Adversarial Networks} &\multirow{2}{*}{\shortstack[c]{Deep\\fakes}} & \multicolumn{2}{c}{Low level vision} & \multicolumn{2}{c}{Perceptual loss} &\multirow{2}{*}{Guided} & \multicolumn{3}{c}{LDM} & \multicolumn{3}{c}{Glide} & \multirow{2}{*}{DALL-E} & Total \\
				\cmidrule(lr){2-7} \cmidrule(lr){9-10} \cmidrule(lr){11-12} \cmidrule(lr){14-16} \cmidrule(lr){17-19} \cmidrule(lr){21-21}

				& \shortstack[c]{Pro-\\GAN} & \shortstack[c]{Cycle-\\GAN} & \shortstack[c]{Big-\\GAN} & \shortstack[c]{Style-\\GAN} & \shortstack[c]{Gau-\\GAN} &  \shortstack[c]{Star-\\GAN}   &  & SITD & SAN & CRN & IMLE & & \shortstack[c]{200\\steps} & \shortstack[c]{200\\w/ CFG} & \shortstack[c]{100\\steps} & \shortstack[c]{100\\27} & \shortstack[c]{50\\27} & \shortstack[c]{100\\10} & & \shortstack[c]{Avg.\\Acc.(\%)}
				\\ 
				\midrule	
				
				{\shortstack[c]{CNNDet}}~\cite{wang2020cnn}  & 99.99 & 85.20 & 70.20 & 85.70 & 78.95 & 91.70 & 53.47 & 66.67 & 48.69 & 86.31 & 86.26 & 60.07 & 54.03 & 54.96 & 54.14 & 60.78 & 63.80 & 65.66 & 55.58 & 69.58 \\

				{\shortstack[c]{Patchfor}}~\cite{chai2020makes}   &  75.03 & 68.97 & 68.47 & 79.16 & 64.23 & 63.94 & 75.54 & \textbf{75.14} & \textbf{75.28} & 72.33 & 55.30 & 67.41 & 76.50 & 76.10 & 75.77 & 74.81 & 73.28 & 68.52 & 67.91 & 71.24\\
				
				

				{\shortstack[c]{Co-occurence}}~\cite{nataraj2019detecting} & 97.70 & 63.15 & 53.75 & 92.50 & 51.10 & 54.70 & 57.10 & 63.06 & 55.85 & 65.65 & 65.80 & 60.50 & 70.70 & 70.55 & 71.00 & 70.25 & 69.60 & 69.90 & 67.55 & 66.86\\

				\shortstack[c]{Spec}~\cite{zhang2019detecting} & 49.90 & \textbf{99.90} & 50.50 & 49.90 & 50.30 & \textbf{99.70} & 50.10 & 50.00 & 48.00 & 50.60 & 50.10 & 50.90 & 50.40 & 50.40 & 50.30 & 51.70 & 51.40 & 50.40 & 50.00 & 55.45\\

				\shortstack[c]{DIRE$^{*}$}~\cite{wang2023dire} & 100.0 & 67.73 & 64.78 & 83.08 & 65.30 & 100.0 & 94.75& 57.62 & 60.96 & 62.36& 62.31 & 83.20& 82.70& 84.05& 84.25& 87.10 &90.80 & 90.25 & 58.75& 77.89\\
				
				{\shortstack[c]{{Ojha~\etal$^{*}$}}}~\cite{ojha2023towards}  & \textbf{100.0} & 98.50 & {94.50} & 82.00 & \textbf{99.50} & 97.00 & 66.60 & 63.00 & 57.50 & 59.50 & 72.00 & 70.03 & {94.19} & 73.76 & {94.36} & 79.07 & 79.85 & 78.14 & {86.78} & 81.38 \\ \hline
				
				{\shortstack[c]{{Teacher}}}  & \textbf{100.0} & {88.05} & {93.40} & {87.55}& {99.00} &84.40 & {74.50} & 71.50 & {75.00} & 86.90 & 95.15 & {83.40} & {95.20} & {83.45} & {95.15} & {91.70} &{91.35} &{89.70} & 
				{85.60} & {87.95}\\		
				
				{\shortstack[c]{{Teacher-Student AD}}}  & {98.50} & {99.00} & {98.75} & {98.80}& {98.75} & 94.65 & {81.85} & 64.50 & 53.50 & 85.85 & 97.55 & {97.40} & {97.25} & \textbf{98.70} & \textbf{98.75} & \textbf{98.75} &{98.70} &{98.70} & {95.75} & {92.41}\\	
				
				{\shortstack[c]{{Teacher-Student AD w/fake}}}  & {98.50} & {98.95} & {98.75}& {98.85} & 98.75 & {93.35} & {86.75} & 57.00 & 55.50 & \textbf{96.70} & \textbf{98.75} & {98.35} & {98.75} & {98.40} & \textbf{98.75} &\textbf{98.75}  &\textbf{98.75} & \textbf{98.75} &97.50 &93.15\\	
				
				{\shortstack[c]{{GenDet}}}  & {99.00} & 99.50 & \textbf{99.30} & \textbf{99.05} & {99.00} & 96.75 & \textbf{88.20} & 63.50 & 67.50 & {93.90} & \textbf{98.75} & \textbf{98.70} & \textbf{98.80} & {98.60} & \textbf{98.75} & \textbf{98.75} & \textbf{98.75} & \textbf{98.75} & \textbf{98.45} & \textbf{94.42}\\		
				
				\bottomrule
		\end{tabular}}
	}
	\caption{Results on UniversalFakeDetect dataset. The methods are compared with the average accuracy as an evaluation metric.}
	\label{tab:Performance_Comparison_on_UniversalFakeDetect_acc}
\end{table*}

\subsection{GenDet on UniversalFakeDetect Dataset}
\paragraph{Comparision to existing methods.} On the UniversalFakeDetect dataset, we benchmark our approach against prior detectors, underscoring the superior performance of our detector, especially when contending with unseen generators. While the majority of methods are trained on fake images generated by ProGAN~\cite{karras2017progressive} in the experiment, Spec~\cite{zhang2019detecting} diverges by employing CycleGAN~\cite{zhu2017unpaired}. The test set, in contrast, encompasses synthetic images produced by a series of various generators. Table~\ref{tab:Performance_Comparison_on_UniversalFakeDetect_ap} and Table~\ref{tab:Performance_Comparison_on_UniversalFakeDetect_acc} show the results of various methods on the UniversalFakeDetect dataset using mAP and average accuracy as evaluation metrics. These methods include representative recent work such as CNNDet~\cite{wang2020cnn}, Patchfor~\cite{chai2020makes}, co-occurrence~\cite{nataraj2019detecting}, Spec~\cite{zhang2019detecting}, DIRE, and Ojha~\etal~\cite{ojha2023towards}. We notice that all of these methods are highly accurate on the same generator and can almost completely identify all real and fake images. However, other generators all suffer from different degrees of degradation, e.g., CNNDet has an AP of 100\% on ProGAN but 59.47\% on SAN. While prior methods exhibit a degree of generalization, there remains potential for further enhancement. CNNDet~\cite{wang2020cnn} is using data augmentation to mitigate the problem of degradation in cross-generator detection. Other methodologies, including patch-level classification~\cite{chai2020makes}, leveraging co-occurrence matrices~\cite{nataraj2019detecting}, or using the frequency space~\cite{zhang2019detecting}, fall short in adequately addressing this challenge. Ojha~\etal~\cite{ojha2023towards} utilizes a CLIP model that has been pre-trained on internet-scale image-text pairs to extract features, and then KNN or linear probes are used for classification. The outcomes from this approach demonstrate the aptitude of the pre-trained model for this task. Therefore, it is feasible to exploit the CLIP and propose a more effective generalization mechanism. We can observe that our method transcends all previous approaches in this way of thinking. In particular, we achieve 94.42\% average accuracy and 95.64\% mAP. This outperforms the existing Ojha~\etal~\cite{ojha2023towards} with 81.38\% average accuracy and 93.38\% mAP, demonstrating the benefits of introducing our proposed adversarial teacher-student discrepancy-aware framework. The advantage mainly comes from the focus on amplifying the output discrepancy of the detector when fed with real and fake images.

\begin{table*}[htb]
	\centering
	\caption{Comparison of GenDet and the other methods. The models are trained on SD V1.4 and evaluated on different testing subsets. $^{*}$ denotes our implementation based on the official codes.}
	\setlength{\tabcolsep}{1.0mm}{
		\begin{tabular}{c|cccccccc|cc}
			\hline
			\multicolumn{1}{c|}{\multirow{2}{*}{Method}} & \multicolumn{8}{c|}{Testing Subset}                                               & \multicolumn{1}{c}{\multirow{2}{*}{\begin{tabular}[c]{@{}c@{}}Avg \\      Acc.(\%)\end{tabular}}} \\ \cline{2-9}
			\multicolumn{1}{c|}{}  & Midjourney & SD V1.4  & SD V1.5  & ADM     & GLIDE   & Wukong  & VQDM    & BigGAN  & \multicolumn{1}{c}{}                                                                           \\ \hline
			ResNet-50~\cite{he2016deep}                                     & 54.9       & \textbf{99.9}     & 99.7    & {53.5}    & 61.9   & 98.2   & 56.6   & 52.0   & 72.1                                                                                       \\
			DeiT-S~\cite{touvron2021training}                                      & 55.6     & \textbf{99.9}   & {99.8}  & 49.8  & 58.1  & 98.9  & 56.9  & 53.5  & 71.6                                                                                        \\
			Swin-T~\cite{liu2021swin}                                      & {62.1}     & \textbf{99.9}   & {99.8}   & 49.8  & {67.6}  & 99.1  & {62.3}  & {57.6}  & {74.8}                                                                                        \\				
			CNNDet~\cite{wang2020cnn}                              & 52.8     & 96.3   & 95.9  & 50.1 & 39.8 & 78.6 & 53.4 & 46.8 & 64.2                                                                                        \\
			Spec~\cite{zhang2019detecting}                                  & 52.0      & 99.4     & 99.2  & 49.7  & 49.8 & 94.8  & 55.6  & 49.8 & 68.8                                                                                        \\
			F3Net~\cite{qian2020thinking}                                & 50.1     & \textbf{99.9} & \textbf{99.9} & 49.9 & 50.0 & \textbf{99.9} & 49.9 & 49.9 & 68.7                                                                                        \\
			GramNet~\cite{liu2020global}                             & 54.2    & 99.2   & 99.1  & 50.3& 54.6  & 98.9  & 50.8 & 51.7  & 69.9                                                                                        \\ 	
			DIRE$^{*}$~\cite{wang2023dire}                             & 60.2    & \textbf{99.9}   & {99.8}  & 50.9 & 55.0  & 99.2  & 50.1 & 50.2  & 70.7                                                                                        \\
			Ojha~\etal$^{*}$~\cite{ojha2023towards}			& 73.2    & 84.2   & 84.0  & {55.2} & 76.9  & 75.6  & 56.9 & \textbf{80.3} & 73.3                                                                                        \\
			
			GenDet                             & \textbf{89.6}    & {96.1}   & {96.1}  & \textbf{58.0} & \textbf{78.4}  & {92.8}  & \textbf{66.5} & {75.0}  & \textbf{81.6}                                                                                        \\				
			\hline
		\end{tabular}
		
	}
\label{tab:Performance_Comparison_on_GenDet}
\end{table*}

\begin{table*}[tb]
	\caption{Model evaluation on degraded images. q is quality. The models are trained and evaluated on SD V1.4. $^{*}$ denotes our implementation based on the official codes.}
	\centering
	\label{tab:perturbed_images}
	\setlength{\tabcolsep}{1.5mm}{
		\begin{tabular}{cccccccc}
			\hline
			\multicolumn{1}{c|}{\multirow{2}{*}{Method}} & \multicolumn{6}{c|}{Testing Subset}       & \multicolumn{1}{c}{\multirow{2}{*}{\begin{tabular}[c]{@{}c@{}}Avg\\ Acc.(\%)\end{tabular}}} \\ \cline{2-7}
			\multicolumn{1}{c|}{}   & LR (112) & LR (64) & JPEG (q=65) & JPEG (q=30) & Blur ($\sigma$=3) & \multicolumn{1}{c|}{Blur ($\sigma$=5)} & \multicolumn{1}{c}{}                                                                       \\ \hline
			\multicolumn{1}{l|}{ResNet-50~\cite{he2016deep}}                & 96.2     & 57.4    & 51.9        & 51.2        & \textbf{97.9}              & \multicolumn{1}{c|}{69.4}              & 70.6                                                                                       \\
			\multicolumn{1}{l|}{DeiT-S~\cite{touvron2021training}}                  & 97.1     & 54.0      & 55.6        & 50.5        & 94.4              & \multicolumn{1}{c|}{67.2}              & 69.8                                                                                       \\
			\multicolumn{1}{l|}{Swin-T~\cite{liu2021swin} }                  & {97.4}     & 54.6    & 52.5        & 50.9        & 94.5              & \multicolumn{1}{c|}{52.5}              & 67.0                                                                                       \\
			\multicolumn{1}{l|}{CNNDet~\cite{wang2020cnn} }         & 50.0       & 50.0      & \textbf{97.3}        & \textbf{97.3}        & {97.4}              & \multicolumn{1}{c|}{{77.9}}              & {78.3}                                                                                       \\
			\multicolumn{1}{l|}{Spec~\cite{zhang2019detecting}}             & 50.0       & 49.9    & 50.8        & 50.4        & 49.9              & \multicolumn{1}{c|}{49.9}              & 50.1                                                                                       \\
			\multicolumn{1}{l|}{F3Net~\cite{qian2020thinking} }          & 50.0       & 50.0      & 89.0          & 74.4        & 57.9              & \multicolumn{1}{c|}{51.7}              & 62.1                                                                                       \\
			\multicolumn{1}{l|}{GramNet~\cite{liu2020global}}                  & \textbf{98.8}     & \textbf{94.9}    & 68.8        & 53.4        & 95.9              & \multicolumn{1}{c|}{{81.6}}              & {82.2}                                                                                       \\ 
			\multicolumn{1}{l|}{DIRE$^{*}$~\cite{wang2023dire}}                 & {64.1}     & {53.5}    &85.4        & 65.0        & 88.8              & \multicolumn{1}{c|}{{56.5}}              & {68.9}                                                                                       \\                                   
			\multicolumn{1}{l|}{Ojha~\etal$^{*}$~\cite{ojha2023towards}}                 & {88.2}     & {78.5}    &85.8        & 83.0        & 69.7              & \multicolumn{1}{c|}{{65.7}}              & {78.3}                            
			\\    
			\multicolumn{1}{l|}{GenDet}    & {85.4}  & {84.1} & 94.4       & 94.3       & 84.6      & \multicolumn{1}{c|}{\textbf{82.9}}             & \textbf{87.6}                                                                                       \\ 			
			
			\hline
			&          &         &             &             &                   &                                        &                                                                                           
		\end{tabular}
	}
	\vspace{-4mm}
\end{table*}

%
%
%
%

\begin{table}[htb]
	\centering
	\caption{The impact of hyperparameters $M$ and $\alpha$.}
	\begin{tabular}{c|c|c|c}
		\hline
		$M$   & $\alpha$ & Avg. Acc (\%) & mAP (\%) \\ \hline
		2.0   & 0.01                  &       94.14  & 95.23    \\
		2.0   & 0.1                  &       \textbf{94.42}  & \textbf{95.64}   \\
		2.0   & 1.0                     &      93.31  & 94.71    \\  \hline
		1.0   & 0.1                     &    93.78    & 95.07    \\
		2.0   & 0.1                     &   \textbf{94.42}     & \textbf{95.64}    \\
		4.0   & 0.1                    &    93.72     &  94.96   \\ \hline
	\end{tabular}
	\label{tab:hyperparameter}
	\vspace{-4mm}
\end{table}

\paragraph{Effectiveness of GenDet.} We conduct experiments to demonstrate the effectiveness of each component in GenDet, as shown in Table~\ref{tab:Performance_Comparison_on_UniversalFakeDetect_ap} and Table~\ref{tab:Performance_Comparison_on_UniversalFakeDetect_acc}. On the UniversalFakeDetect dataset, we first train a binary classification model as a teacher with average accuracy and AP as 87.95\% and 93.40\%. This serves as the baseline for our approach, and then we apply the teacher-student anomaly detection method using the teacher model. The teacher network is fixed, and the student network using the same architecture is utilized for training. we only use real images and Equation~\ref{eq:loss_real} to train a student. We use the discrepancy between teacher and student to train a classifier for classification. We find that this improves the performance to average accuracy and AP, respectively. Further, we utilize the fake image input to improve the teacher-student anomaly detection method. The loss function includes Equation~\ref{eq:loss_real} and Equation~\ref{eq:loss_fake}. No feature augmenter is used to augment the fake features in this experimental setting. The binary classifier model is still trained. We find that this improves the performance to average accuracy and AP, respectively. Finally, we add a feature augmenter, and our GenDet improves the average accuracy and mAP to 94.4\% and 95.6\%.

\paragraph{Effectiveness of hyperparameter $M$ and $\alpha$.}
We conduct hyperparameter tuning experiments to verify the effect of adjusting $M$ and $\alpha$ in Equation~\ref{eq:loss_fake}, as shown in Table~\ref{tab:hyperparameter}. $M$ represents the minimum output discrepancy between the teacher and the student in the face of the enhanced fake features. $\alpha$ represents the weight of the loss function in Equation~\ref{eq:loss_fake}. We find that our method is not sensitive to hyperparameters. The margin works best at 2.0, and $\alpha$ works best at 0.1. Therefore, we empirically set the margin to 2.0 and $\alpha$ to 0.1 in the following experiments.

\subsection{GenDet on GenImage Dataset}
\paragraph{Comparision to existing methods.} Compared to the UniversalFakeDetect dataset, GenImage is more meticulously curated and thoroughly evaluated, as shown in Table~\ref{tab:Performance_Comparison_on_GenDet}. The ratio of real and fake images, the number of images generated by each generator, and the quantity of each class of image have all been carefully calibrated in GenImage. It primarily utilizes the Diffusion models as its core generators while also encompassing the GAN family of generators, exemplified by BigGAN. Compared to UniversalFakeDetect, GenImage covers more recent major generators, such as Stable Diffusion and Midjourney, which produce better results and are, therefore, more likely to be used to create fake images that are harmful to society. We compare the previous methods on GenImage. In addition to the methods already evaluated in GenImage, we also re-implement DIRE~\cite{wang2023dire} and Ojha~\etal~\cite{ojha2023towards} based on the official open-source code. It can be seen that the metrics are generally lower for all methods on GenImage, demonstrating that this dataset is more difficult. In this case, our method shows greater advantages compared to the other methods, e.g., we have a higher 8.3\% accuracy than Ojha~\etal~\cite{ojha2023towards}.

\begin{figure*}[t]
	\centering
	\captionsetup[subfigure]{labelformat=empty}
	\begin{subfigure}[b]{0.138\textwidth}
		\centering
		\includegraphics[width=\textwidth]{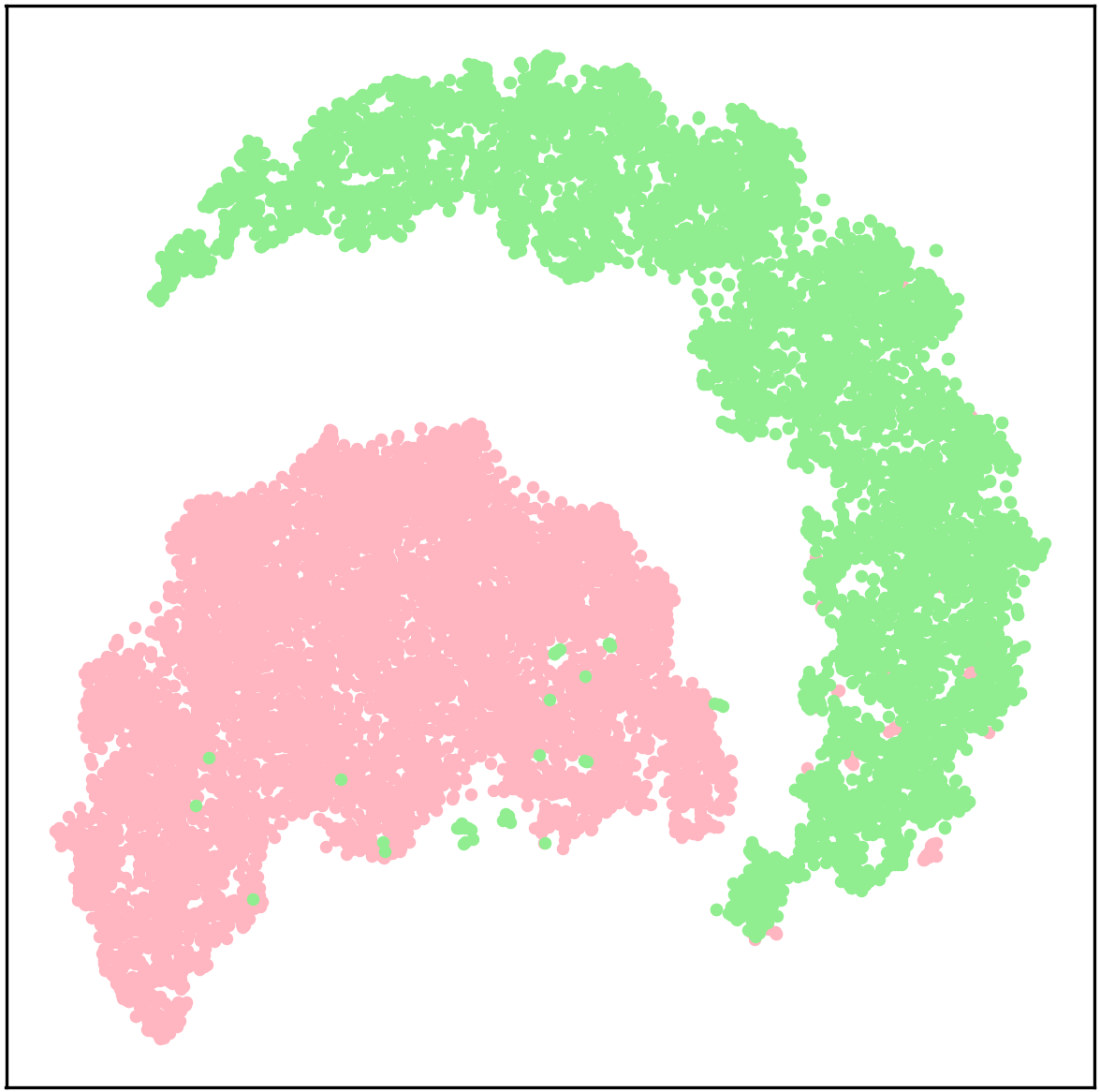}
		\label{fig3:a}
	\end{subfigure}
	\begin{subfigure}[b]{0.138\textwidth}
		\centering
		\includegraphics[width=\textwidth]{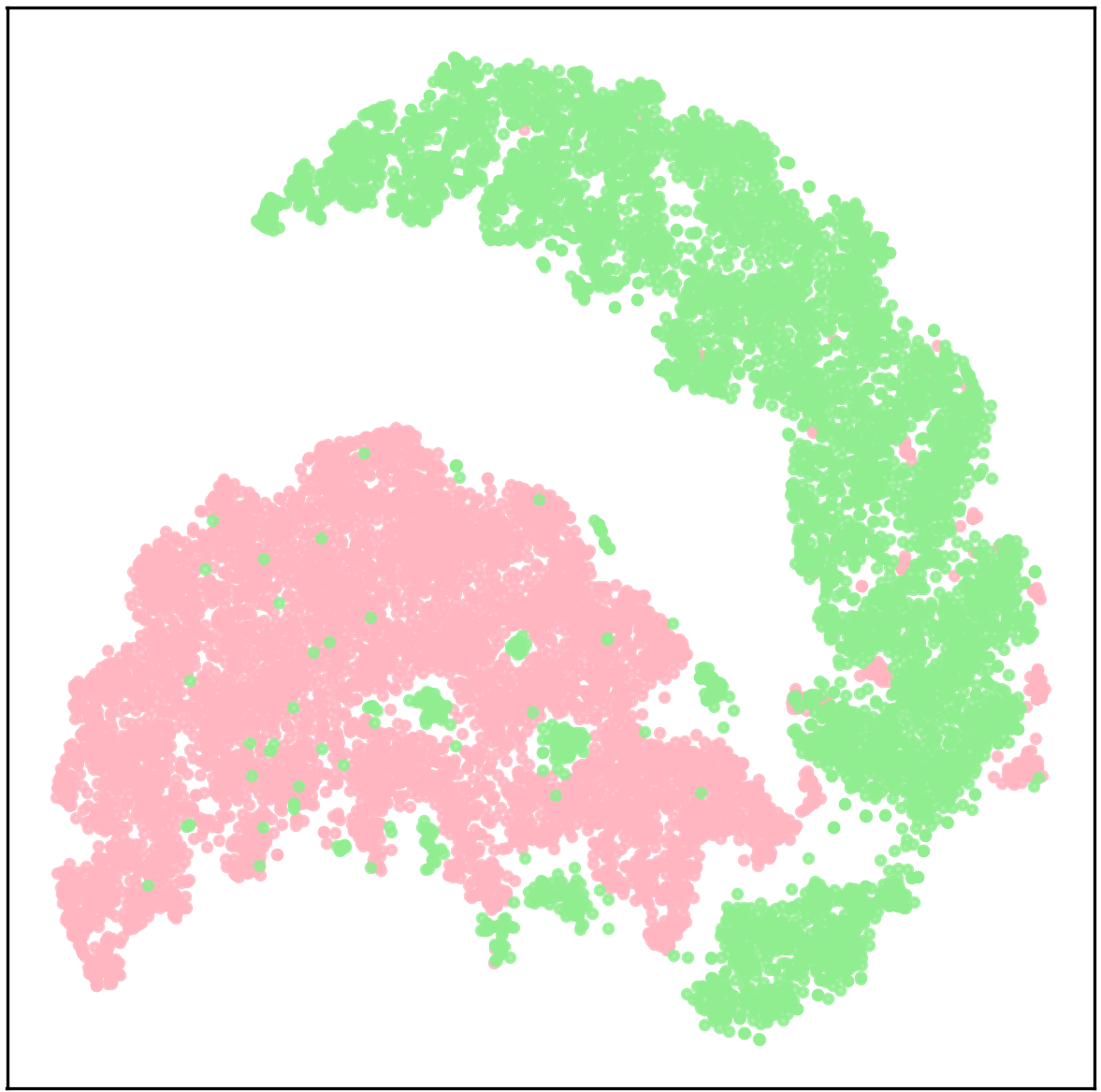}
		\label{fig3:b}
	\end{subfigure}
	\begin{subfigure}[b]{0.138\textwidth}
		\centering
		\includegraphics[width=\textwidth]{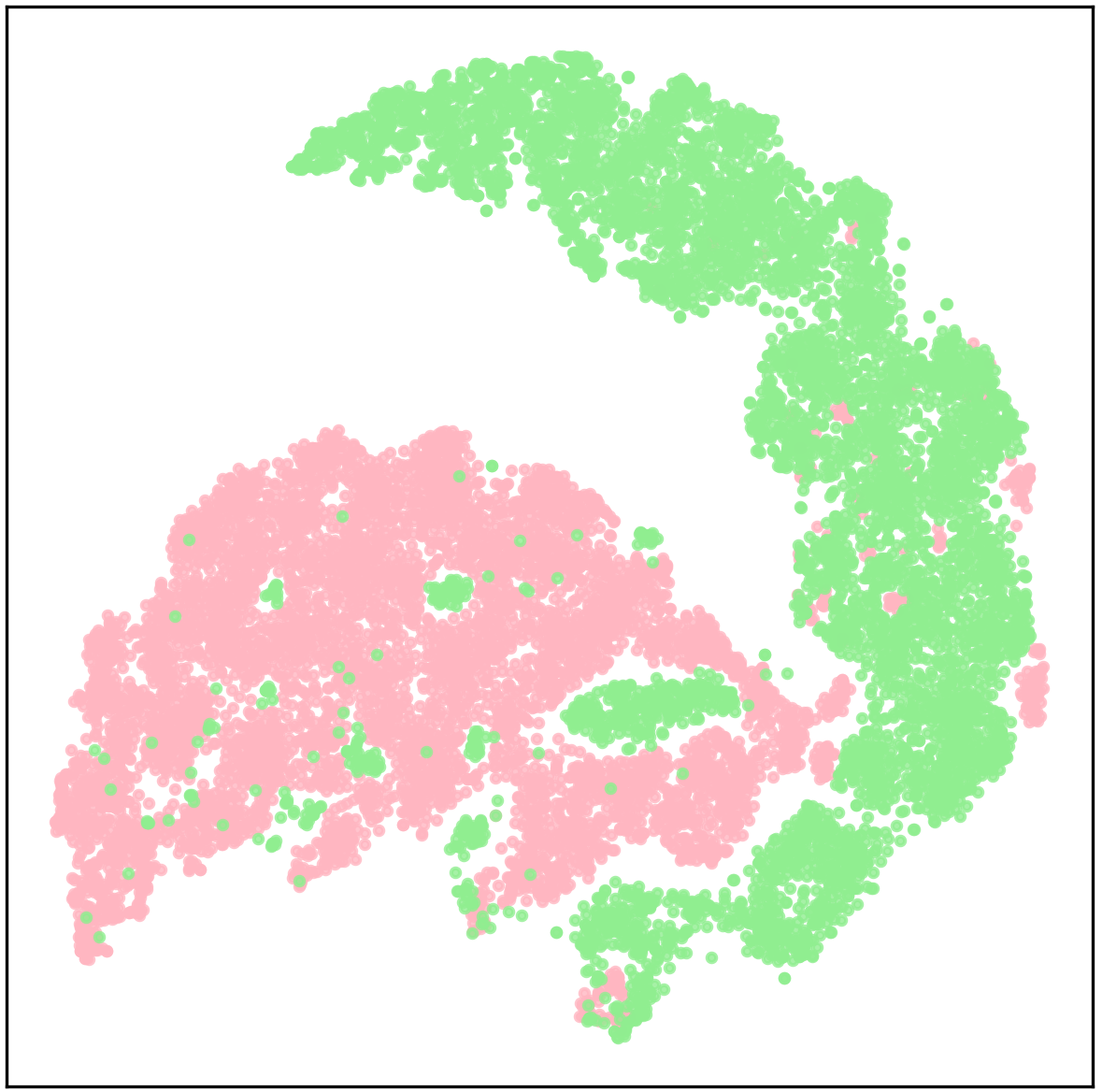}
		\label{fig3:c}
	\end{subfigure}
	\begin{subfigure}[b]{0.138\textwidth}
		\centering
		\includegraphics[width=\textwidth]{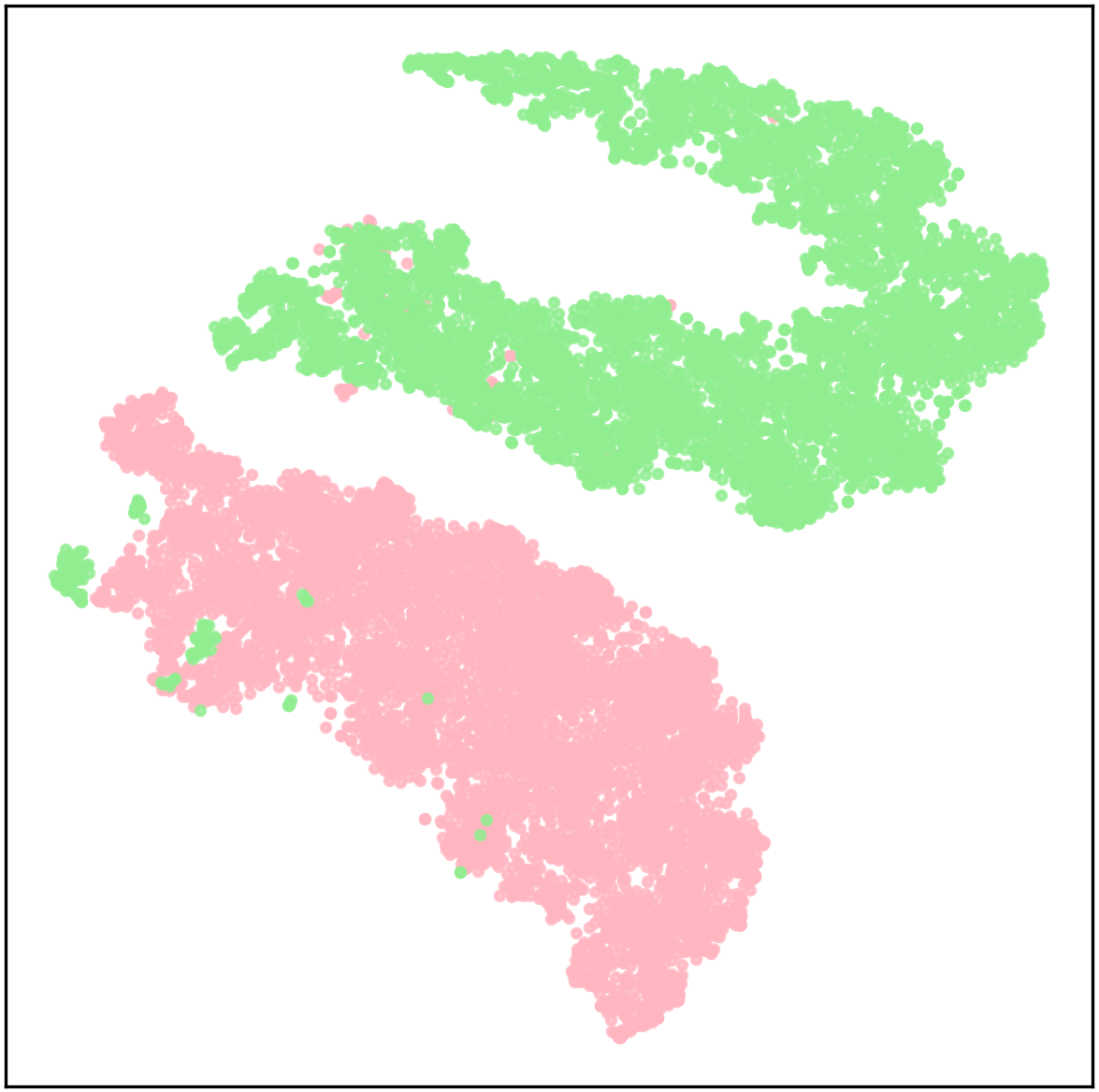}
		\label{fig3:d}
	\end{subfigure}
	\begin{subfigure}[b]{0.138\textwidth}
		\centering
		\includegraphics[width=\textwidth]{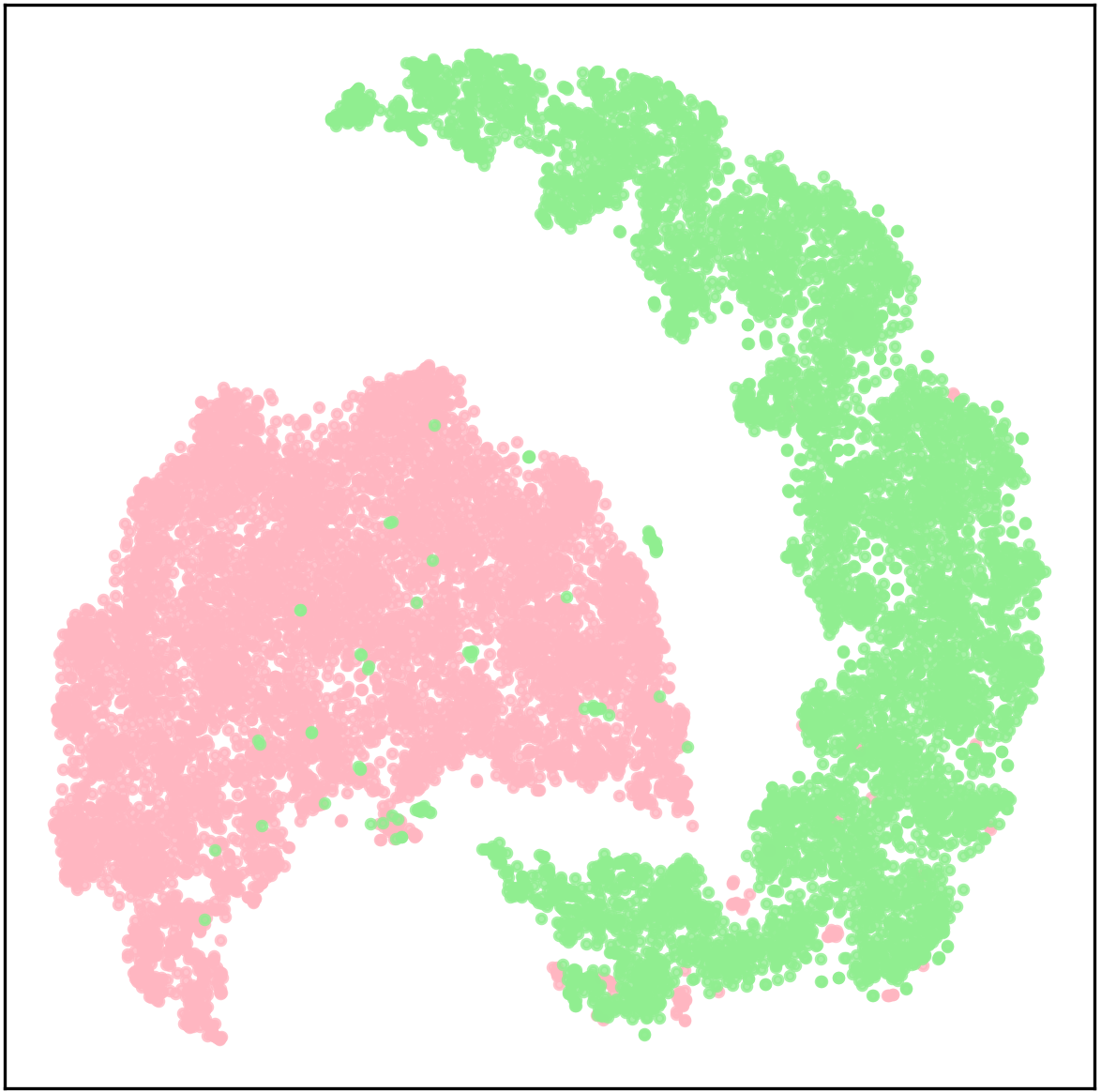}
		\label{fig3:e}
	\end{subfigure}
	\begin{subfigure}[b]{0.138\textwidth}
		\centering
		\includegraphics[width=\textwidth]{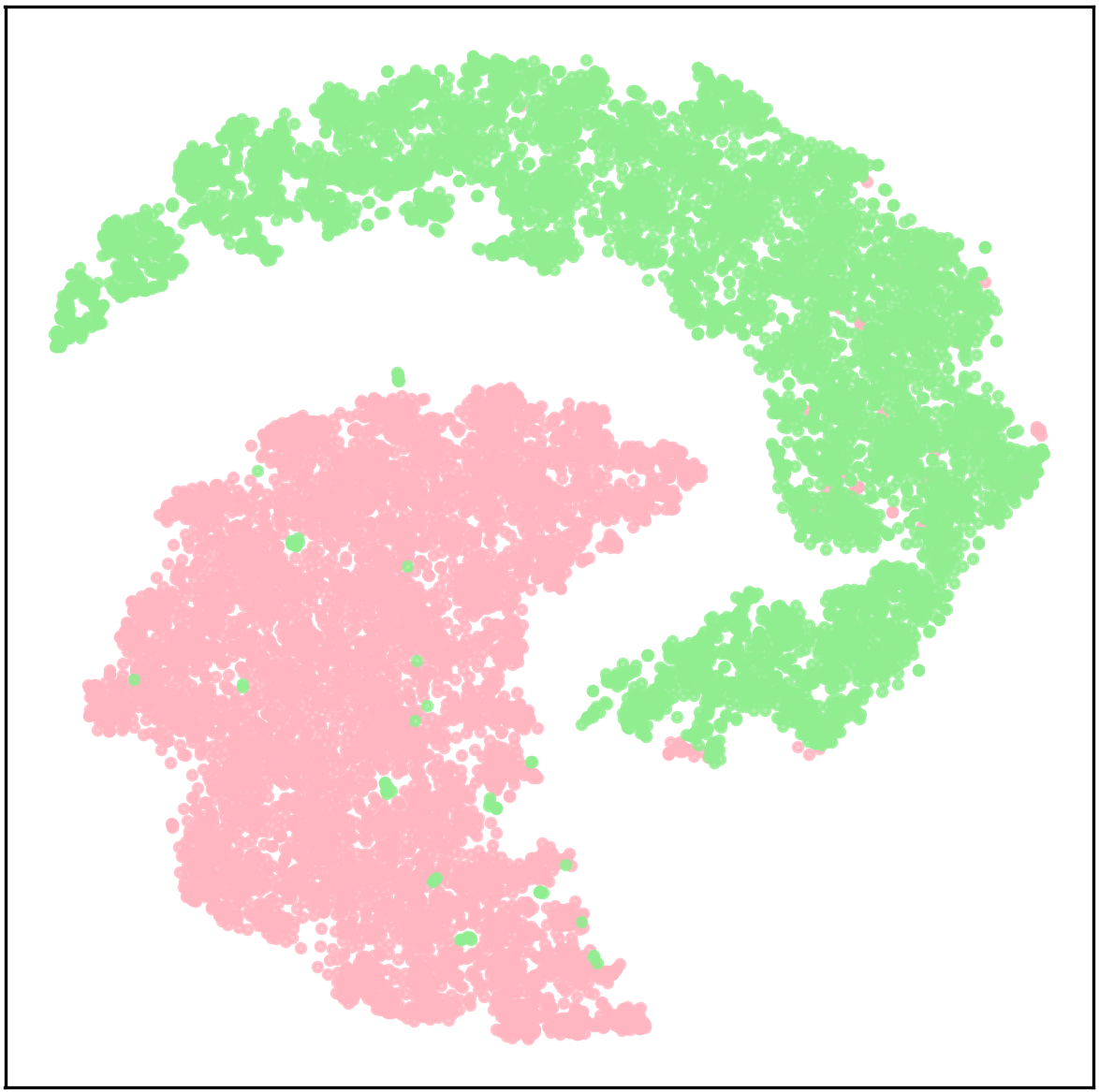}
		\label{fig3:f}
	\end{subfigure}
	\begin{subfigure}[b]{0.138\textwidth}
		\centering
		\includegraphics[width=\textwidth]{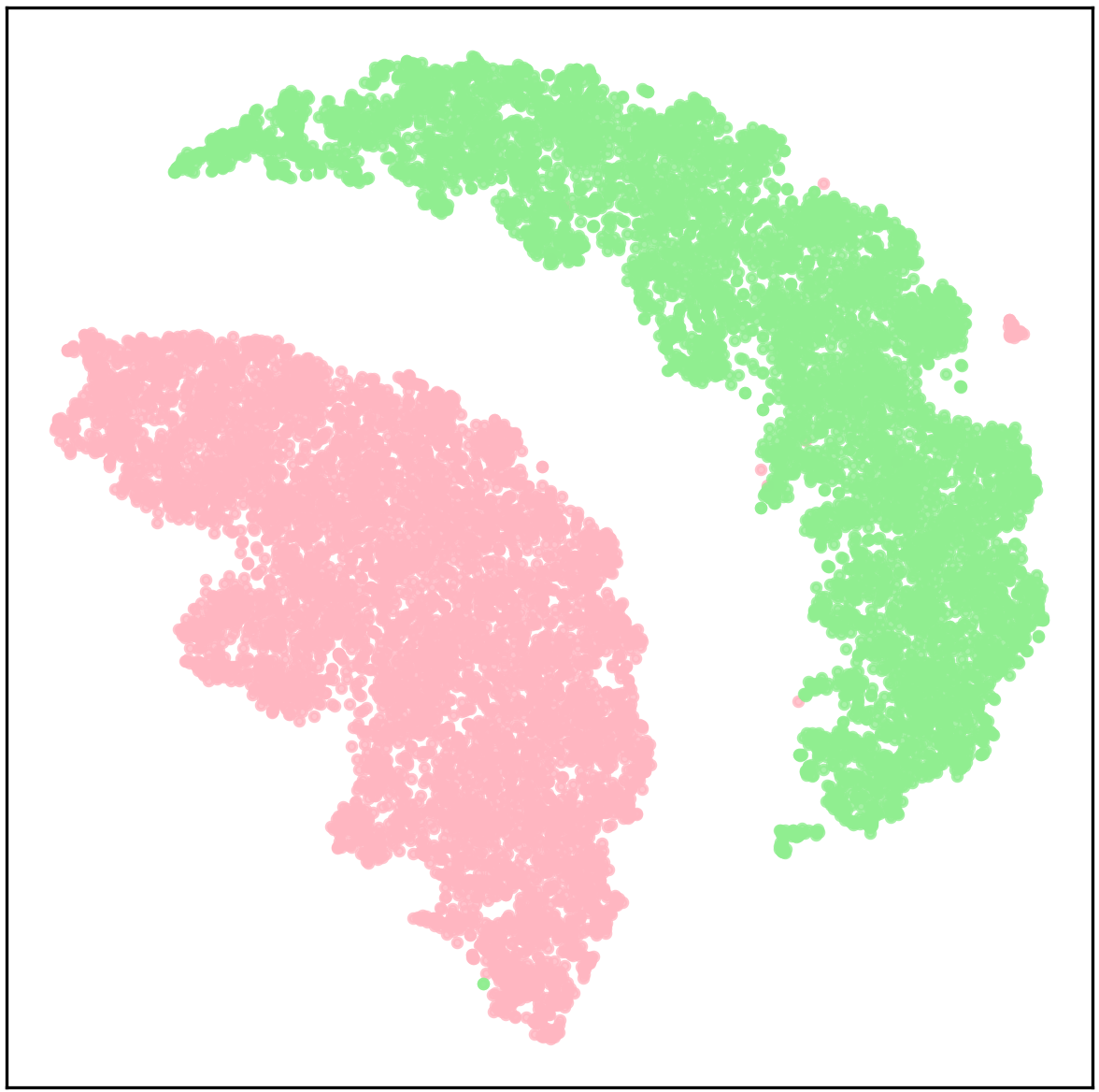}
		\label{fig3:g}
	\end{subfigure}
	\vskip -10pt
	
	\begin{subfigure}[b]{0.138\textwidth}
		\centering
		\includegraphics[width=\textwidth]{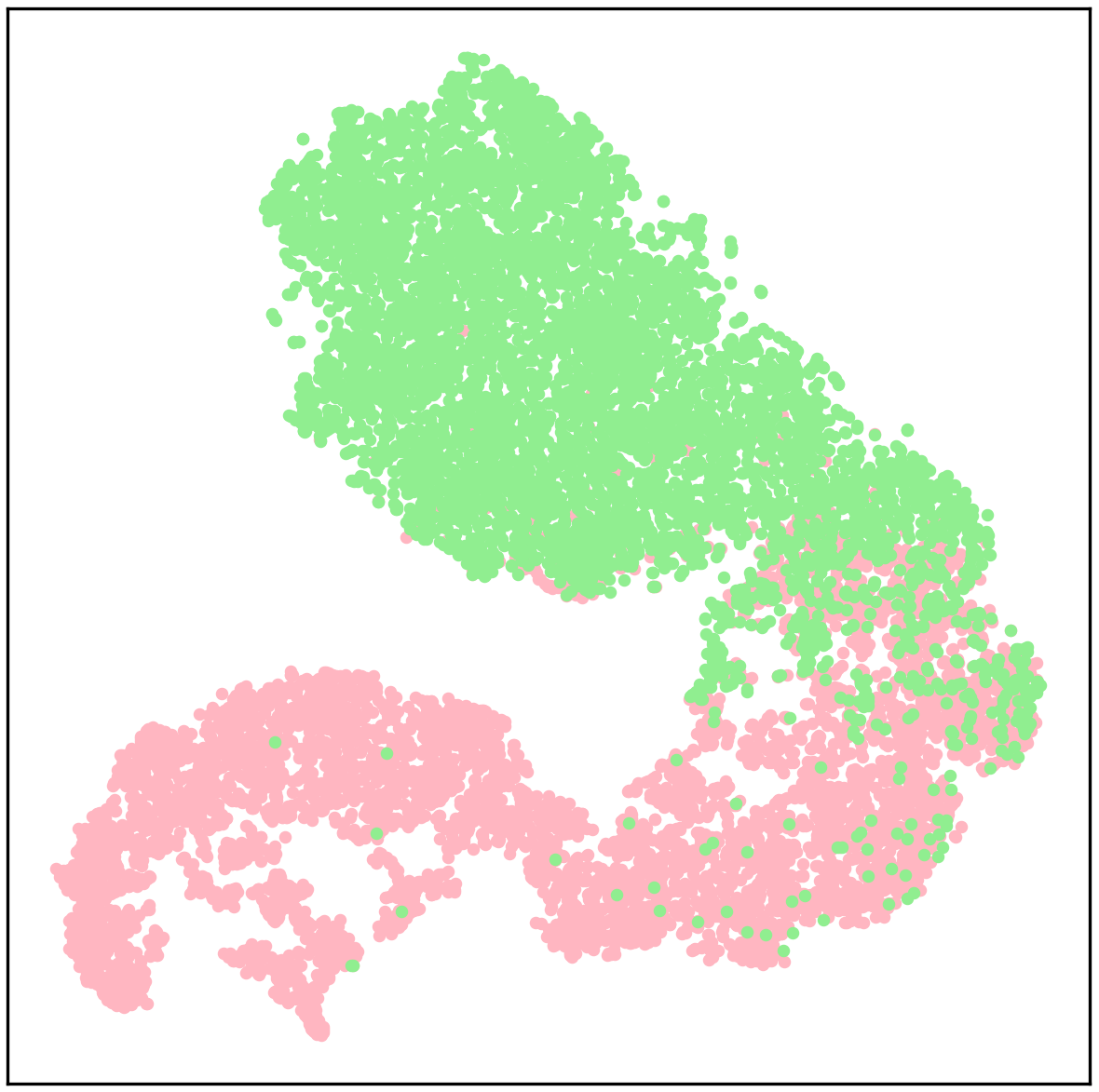}
		\caption{Original}
		\label{fig3:tsne_original_cnndet}
	\end{subfigure}
	\begin{subfigure}[b]{0.138\textwidth}
		\centering
		\includegraphics[width=\textwidth]{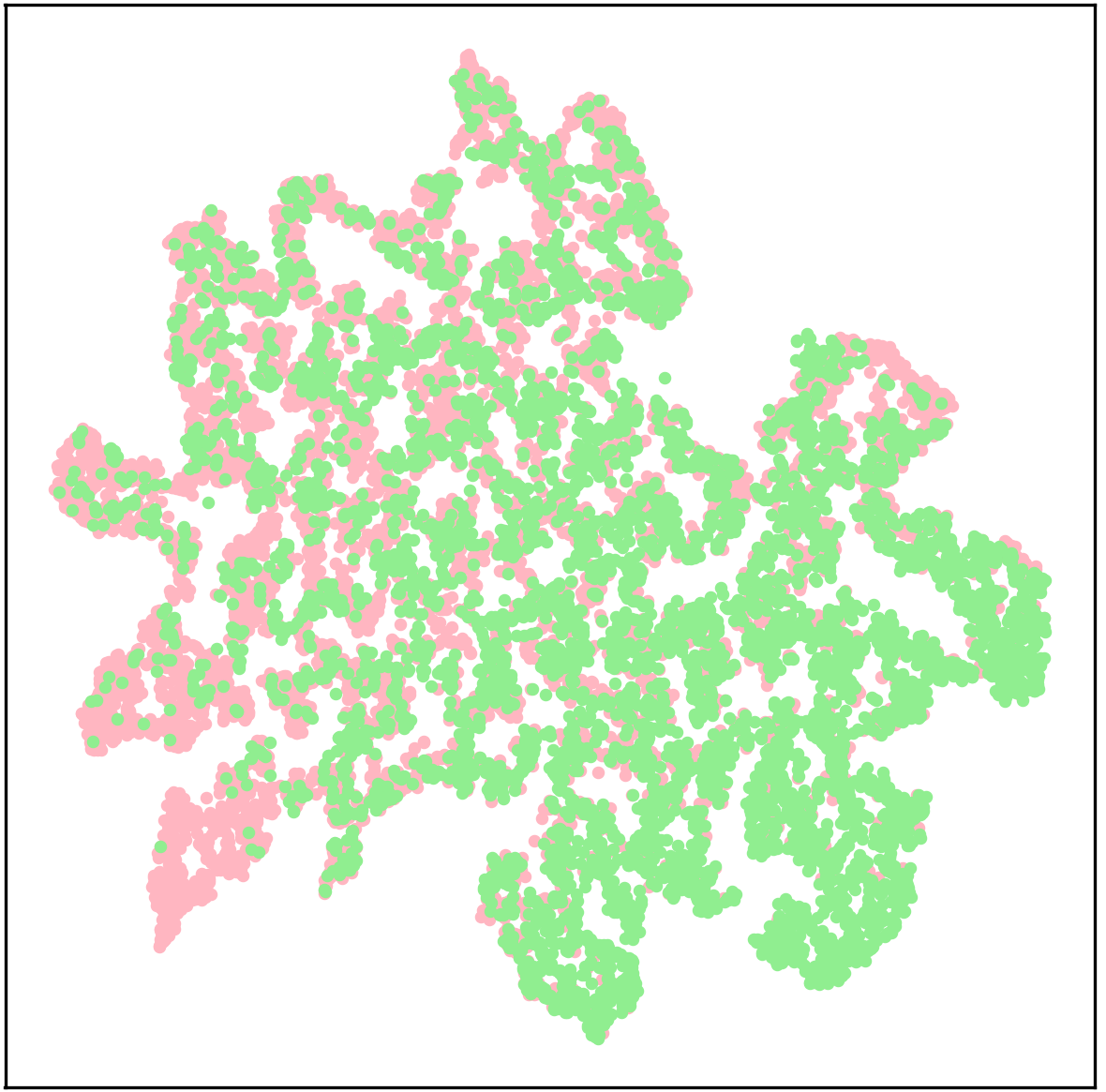}
		\caption{LR (112)}
		\label{fig3:tsne_lr112_cnndet}
	\end{subfigure}
	\begin{subfigure}[b]{0.138\textwidth}
		\centering
		\includegraphics[width=\textwidth]{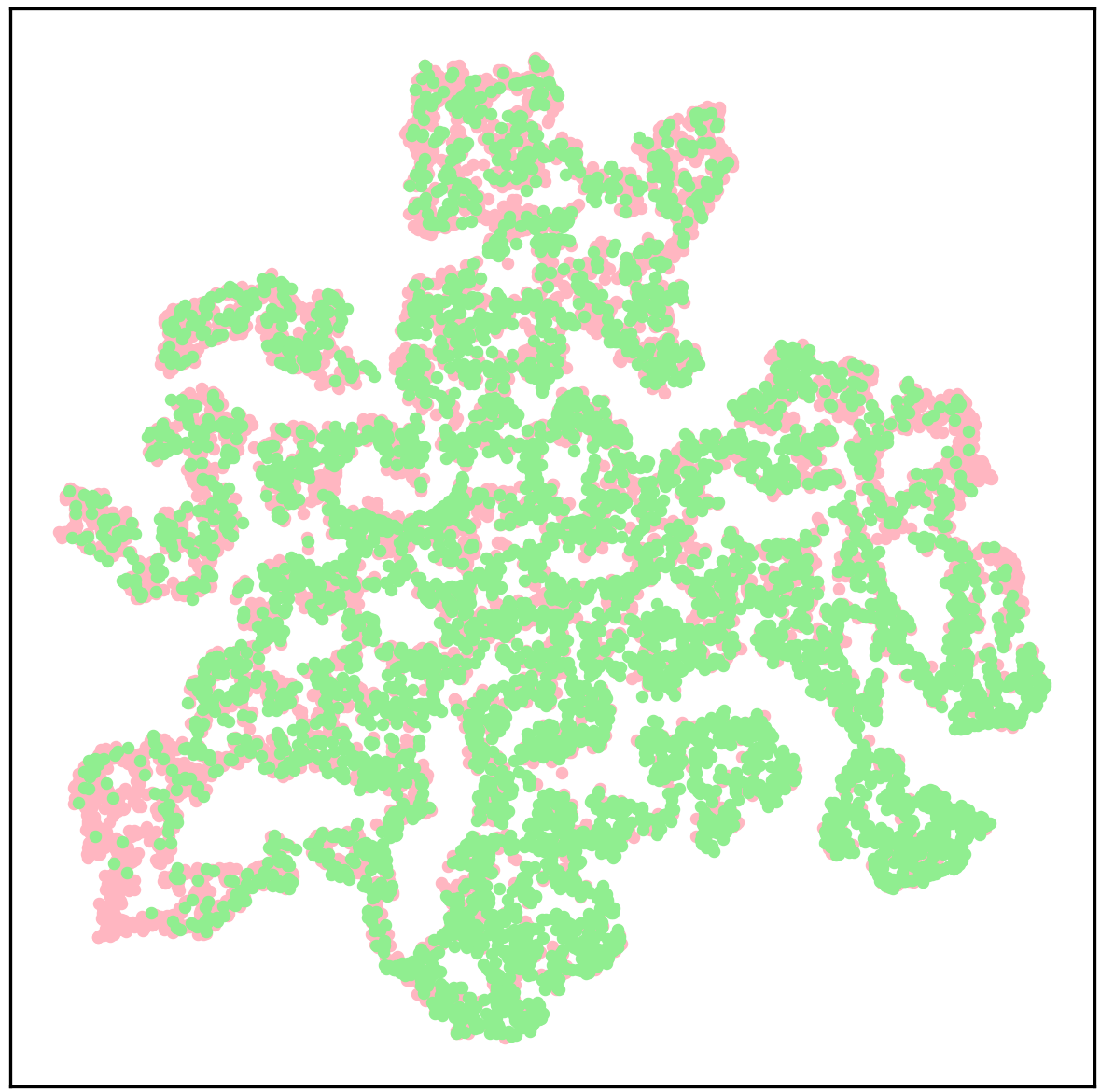}
		\caption{LR (64)}
		\label{fig3:tsne_lr64_cnndet}
	\end{subfigure}
	\begin{subfigure}[b]{0.138\textwidth}
		\centering
		\includegraphics[width=\textwidth]{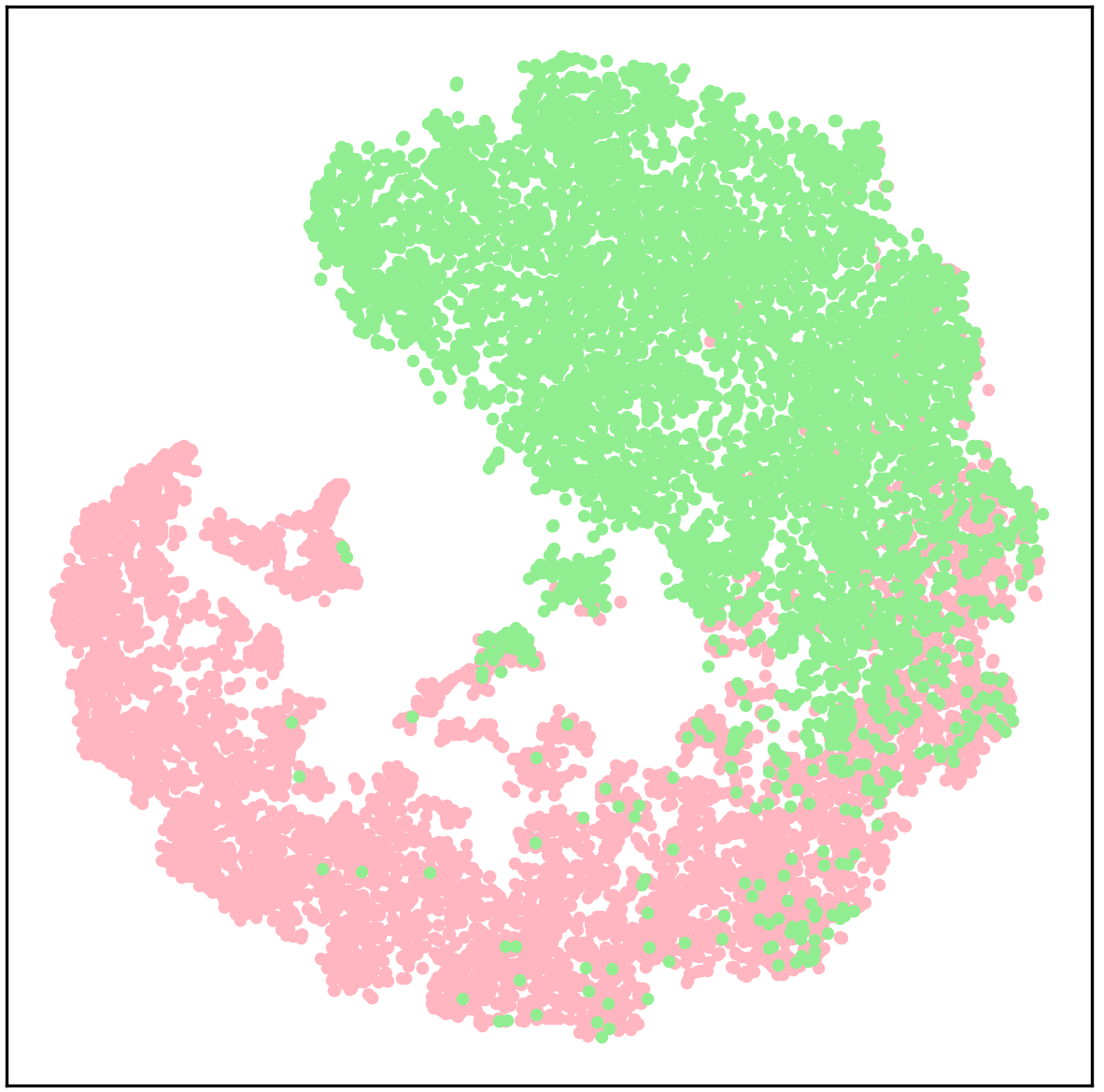}
		\caption{JPEG (q=65)}
		\label{fig3:tsne_jpeg65_cnndet}
	\end{subfigure}
	\begin{subfigure}[b]{0.138\textwidth}
		\centering
		\includegraphics[width=\textwidth]{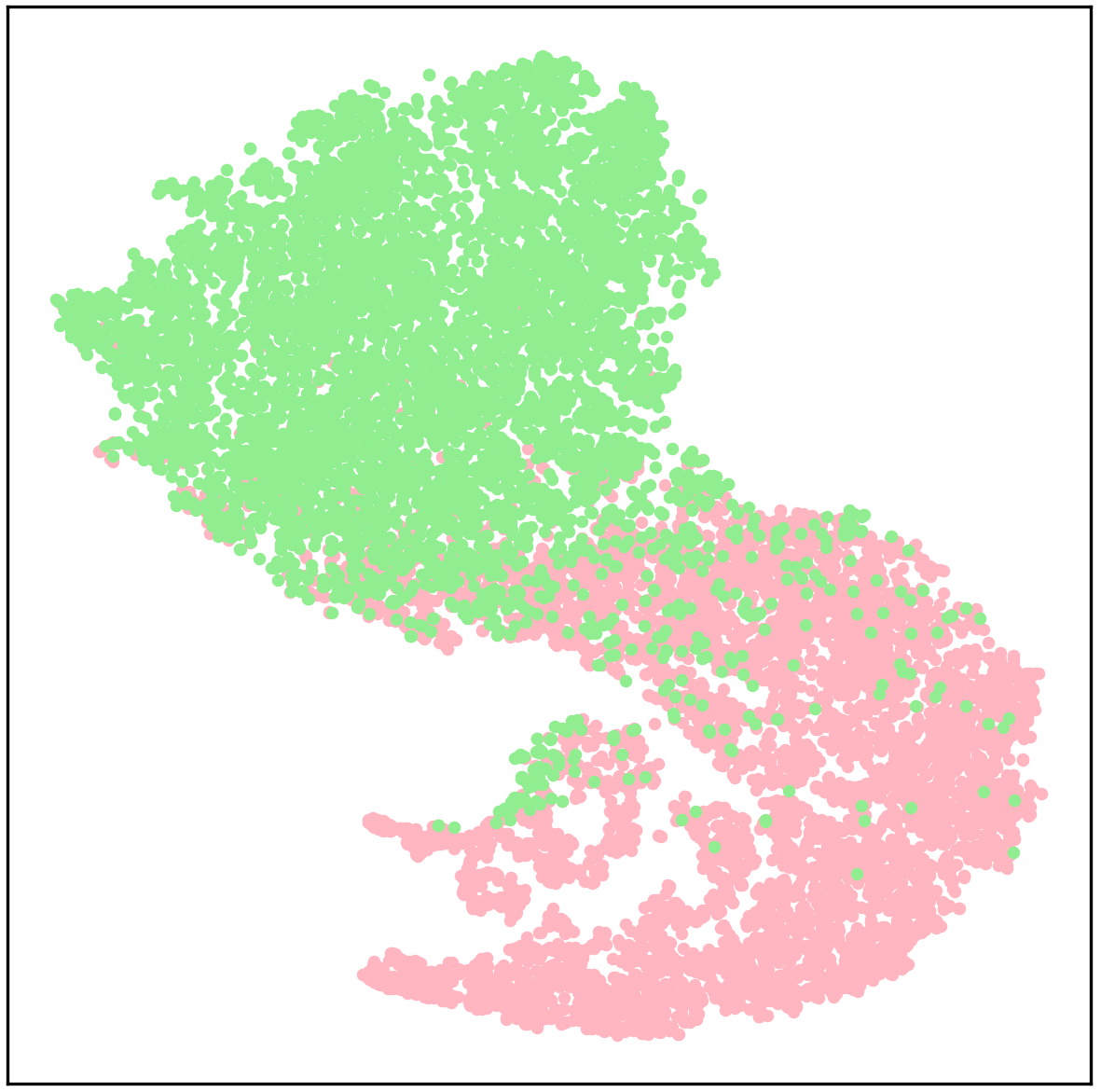}
		\caption{JPEG (q=30)}
		\label{fig3:tsne_jpeg30_cnndet}
	\end{subfigure}
	\begin{subfigure}[b]{0.138\textwidth}
		\centering
		\includegraphics[width=\textwidth]{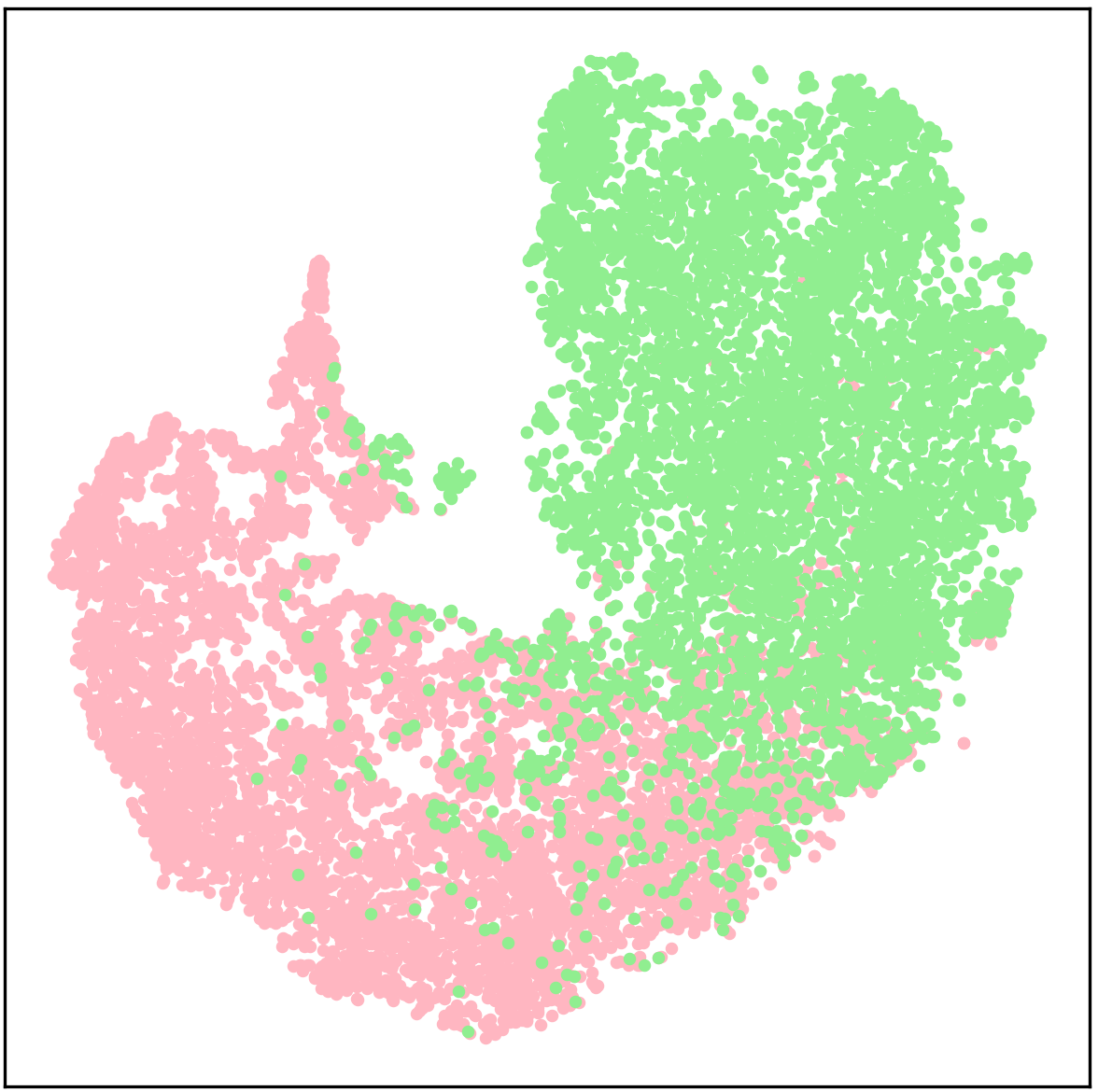}
		\caption{Blur ($\sigma$=3)}
		\label{fig3:tsne_blur3_cnndet}
	\end{subfigure}
	\begin{subfigure}[b]{0.138\textwidth}
		\centering
		\includegraphics[width=\textwidth]{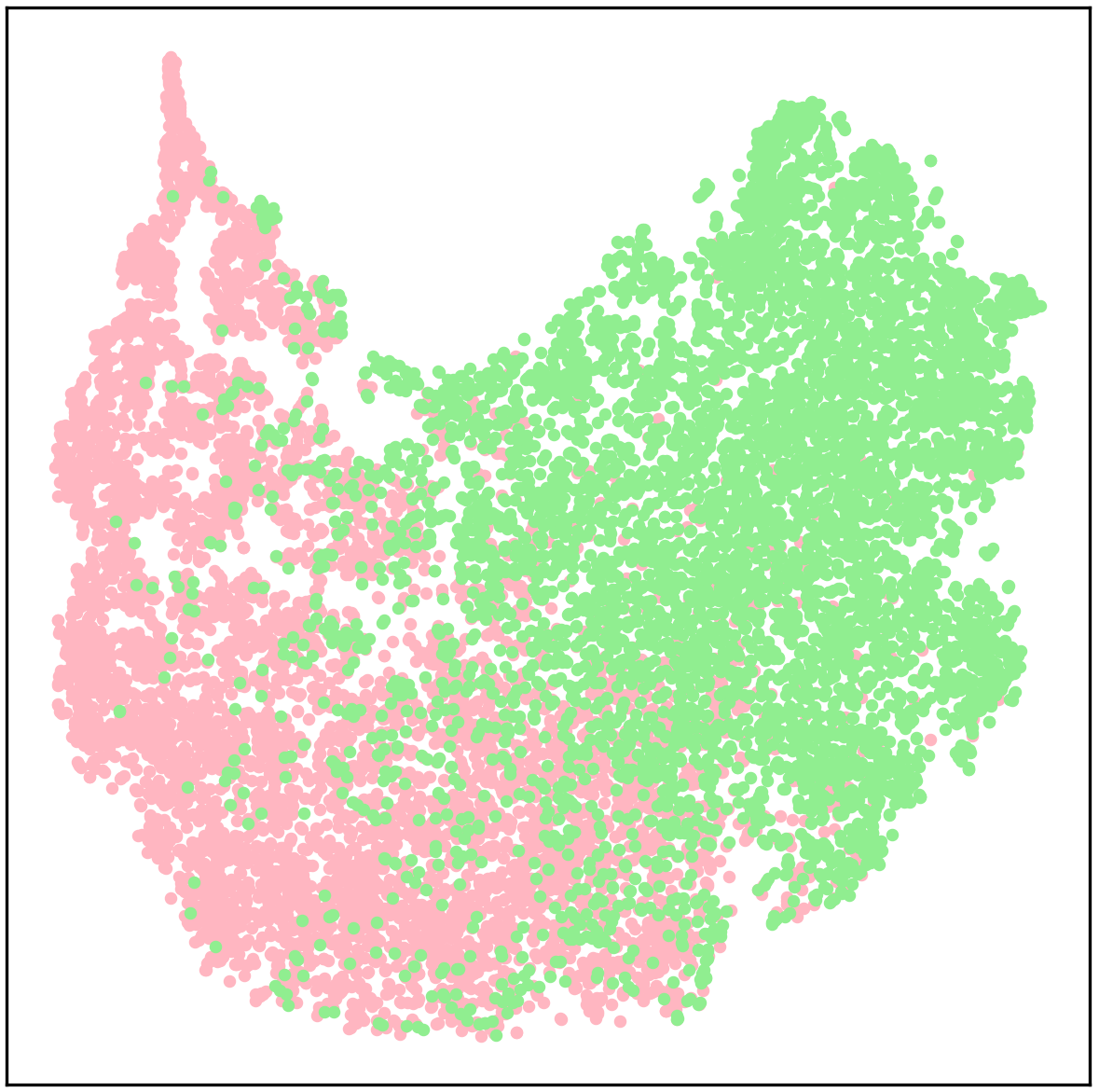}
		\caption{Blur ($\sigma$=5)}
		\label{fig3:tsne_blur5_cnndet}
	\end{subfigure}
	\caption{T-SNE degradation of GenDet (top) and CNNDet (bottom). The real images are shown in red. The fake images are shown in green, respectively. A large distribution discrepancy between real and fake images decreases the difficulty of classification.}
	\label{tsne}
\end{figure*}

\paragraph{Degraded Image Classification.} In the real world, detectors face a wide variety of image degradation problems in addition to cross-generator detection, such as image compression, low resolution, and noise interference. We further examine the performance of detectors on the degraded images following the setting in GenImage, as shown in Table~\ref{tab:perturbed_images}. To evaluate this task, we draw upon the GenImage implementation for post-processing the test images. We train and test the detector on Stable Diffusion V1.4. The images are resized to 112 and 64 at low-resolution settings. We use hyperparameters with quality ratios of 65 and 30 for the JPEG compression case. We use Gaussian blurring to introduce blurring effects. We compare the performance of GenDet with various existing detectors. GramNet achieves the best score (82.2\%) in the existing methods, while our GenDet can achieve 87.6\%. It can be seen that the robustness of our detector is generally better than that of other detectors and can cope well with all types of image degradation problems.

\paragraph{Cross-Dataset Evaluation.} In Table~\ref{tab:cross_dataset_acc} and Table~\ref{tab:cross_dataset_ap}, we perform a cross-dataset evaluation on different kinds of representative generators. We train GenDet on the diffusion model (Stable Diffusion V1.4), and the evaluation is performed on the diffusion model (IF~\cite{IF}), autoregressive model (CogView2~\cite{ding2022cogview2}), and GAN (StyleGAN~\cite{karras2019style}). Following GenImage~\cite{zhu2023genimage}, we use the images from ImageNet and the input sentences to IF and CogView2 following the template "photo of class", with "class" being substituted by ImageNet labels. For each generator, we use 1000 real images and 1000 fake images. The threshold is 0.5. It can be seen that even in the face of a large domain gap, our detector still achieves a detection performance with average acc/mAP of 85.2\% / 95.2\%. This experiment verifies that our proposed detector has good generalization performance.

\begin{table}[htb]
	\small
	\centering
	\caption{Cross-dataset evaluation. The evaluation metric is average accuracy.}
	\begin{tabular}{c|cccc}
		\hline
		& CogView2 & StyleGAN & IF & \begin{tabular}[c]{@{}c@{}}Avg\\ Acc. (\%)\end{tabular}\\ \hline
		ResNet50    &  74.9  &   69.6    &   67.6      &  70.7  \\
		Ojha~\etal$^{*}$     &  75.1  &   44.0    &   80.2      &  66.4  \\
		GenDet      &  85.8  &   81.8    &   87.9      &  85.2  \\ \hline
	\end{tabular}
	\label{tab:cross_dataset_acc}
\end{table}

\begin{table}[htb]
	\small
	\centering
	\caption{Cross-dataset evaluation. The evaluation metric is mAP.}
	
	\begin{tabular}{c|cccc}
		\hline
		& CogView2 & StyleGAN & IF & mAP (\%)\\ \hline
		ResNet50    &  95.9  &   90.0    &   95.3      &  93.7  \\
		Ojha~\etal$^{*}$     &  93.6  &   43.8    &   95.4      &  77.6  \\
		GenDet      &  95.1  &   92.3    &   98.3      &  95.2  \\ \hline
	\end{tabular}
	\label{tab:cross_dataset_ap}
\end{table}

\paragraph{Visulization.} We use t-SNE to visualize the features extracted by our method in the GenImage dataset, as shown in Figure~\ref{tsne}. We refer to Table~\ref{tab:perturbed_images} for degrading images, which are then fed into various methods for comparison. The training and testing images are generated by Stable Diffusion V1.4. We reimplement the CNNDet and achieve the results in Table~\ref{tab:perturbed_images}. We input the images into teacher and student networks and then utilize the discrepancy between the teacher and student for t-SNE visualization. After degradation, the performance of the images decreases to various extents. The original images have not undergone degradation. We can see that the features of real and fake images extracted by the model are well-distinguished. Next, we visualize the features of images that have been downsampled at different ratios. Compared to the original, the feature distributions of these real and fake images become more similar. This is particularly evident when the resolution is reduced from 112 to 64. This distribution trend is also present in JPEG compression and Gaussian Blur. However, overall, even when the input images suffer from image degradation, the features extracted by our model remain easily distinguishable. Therefore, this t-SNE visualization illustrates the robustness of our GenDet method in image degradation compared to other methods.

\section{Conclusion}
\label{sec:conclusion}

In this paper, we introduce GenDet, a novel detector designed to effectively distinguish between real and AI-generated fake images. A key challenge in detecting AI-generated images is identifying images produced by generators not included in the training set. To address this, we propose an adversarial teacher-student discrepancy-aware framework. This framework consists of two parts: teacher-student discrepancy-aware learning and generalized feature augmentation. The former ensures the output of the student network aligns with that of the teacher for real images and diverges for fakes, while the latter uses adversarial learning to train a feature augmenter, promoting smaller output discrepancy with fake images. The generalized feature augmentation enables teacher and student networks trained with a teacher-student discrepancy-aware to exhibit larger discrepancy when encountering images from unseen generators. Experimental results on the GenImage and UniversalFakeDetect datasets demonstrate that our method significantly surpasses existing approaches.

{
    \small
    \bibliographystyle{ieeenat_fullname}
    \bibliography{main}

\begin{thebibliography}{49}
\providecommand{\natexlab}[1]{#1}
\providecommand{\url}[1]{\texttt{#1}}
\expandafter\ifx\csname urlstyle\endcsname\relax
  \providecommand{\doi}[1]{doi: #1}\else
  \providecommand{\doi}{doi: \begingroup \urlstyle{rm}\Url}\fi

\bibitem[bra()]{braintumor}
Brain mri images for brain tumor detection.
\newblock
  \emph{https://www.kaggle.com/datasets/navoneel/brain-mri-images-for-brain-tumor-detection}.

\bibitem[IF(2022)]{IF}
if.
\newblock In \emph{https://github.com/deep-floyd/IF/tree/develop}, 2022.

\bibitem[hem(2022)]{hemorrhage}
Head ct - hemorrhage.
\newblock In
  \emph{https://www.kaggle.com/datasets/felipekitamura/head-ct-hemorrhage},
  2022.

\bibitem[mid(2022)]{midjourney}
Midjourney.
\newblock In \emph{https://www.midjourney.com/home/}, 2022.

\bibitem[wuk(2022)]{wukong}
Wukong.
\newblock In \emph{https://xihe.mindspore.cn/modelzoo/wukong}, 2022.

\bibitem[Bergmann et~al.(2019)Bergmann, Fauser, Sattlegger, and
  Steger]{bergmann2019mvtec}
Paul Bergmann, Michael Fauser, David Sattlegger, and Carsten Steger.
\newblock Mvtec ad--a comprehensive real-world dataset for unsupervised anomaly
  detection.
\newblock In \emph{Proceedings of the IEEE/CVF conference on computer vision
  and pattern recognition}, pages 9592--9600, 2019.

\bibitem[Bergmann et~al.(2020)Bergmann, Fauser, Sattlegger, and
  Steger]{bergmann2020uninformed}
Paul Bergmann, Michael Fauser, David Sattlegger, and Carsten Steger.
\newblock Uninformed students: Student-teacher anomaly detection with
  discriminative latent embeddings.
\newblock In \emph{Proceedings of the IEEE/CVF conference on computer vision
  and pattern recognition}, pages 4183--4192, 2020.

\bibitem[Bird and Lotfi(2023)]{bird2023cifake}
Jordan~J Bird and Ahmad Lotfi.
\newblock Cifake: Image classification and explainable identification of
  ai-generated synthetic images.
\newblock \emph{arXiv preprint arXiv:2303.14126}, 2023.

\bibitem[Brock et~al.(2018)Brock, Donahue, and Simonyan]{brock2018large}
Andrew Brock, Jeff Donahue, and Karen Simonyan.
\newblock Large scale gan training for high fidelity natural image synthesis.
\newblock \emph{arXiv preprint arXiv:1809.11096}, 2018.

\bibitem[Chai et~al.(2020)Chai, Bau, Lim, and Isola]{chai2020makes}
Lucy Chai, David Bau, Ser-Nam Lim, and Phillip Isola.
\newblock What makes fake images detectable? understanding properties that
  generalize.
\newblock In \emph{Computer Vision--ECCV 2020: 16th European Conference,
  Glasgow, UK, August 23--28, 2020, Proceedings, Part XXVI 16}, pages 103--120.
  Springer, 2020.

\bibitem[Chen et~al.(2018)Chen, Chen, Xu, and Koltun]{chen2018learning}
Chen Chen, Qifeng Chen, Jia Xu, and Vladlen Koltun.
\newblock Learning to see in the dark.
\newblock In \emph{Proceedings of the IEEE conference on computer vision and
  pattern recognition}, pages 3291--3300, 2018.

\bibitem[Chen and Koltun(2017)]{chen2017photographic}
Qifeng Chen and Vladlen Koltun.
\newblock Photographic image synthesis with cascaded refinement networks.
\newblock In \emph{Proceedings of the IEEE international conference on computer
  vision}, pages 1511--1520, 2017.

\bibitem[Choi et~al.(2018)Choi, Choi, Kim, Ha, Kim, and Choo]{choi2018stargan}
Yunjey Choi, Minje Choi, Munyoung Kim, Jung-Woo Ha, Sunghun Kim, and Jaegul
  Choo.
\newblock Stargan: Unified generative adversarial networks for multi-domain
  image-to-image translation.
\newblock In \emph{Proceedings of the IEEE conference on computer vision and
  pattern recognition}, pages 8789--8797, 2018.

\bibitem[Dai et~al.(2019)Dai, Cai, Zhang, Xia, and Zhang]{dai2019second}
Tao Dai, Jianrui Cai, Yongbing Zhang, Shu-Tao Xia, and Lei Zhang.
\newblock Second-order attention network for single image super-resolution.
\newblock In \emph{Proceedings of the IEEE/CVF conference on computer vision
  and pattern recognition}, pages 11065--11074, 2019.

\bibitem[Deng and Li(2022)]{deng2022anomaly}
Hanqiu Deng and Xingyu Li.
\newblock Anomaly detection via reverse distillation from one-class embedding.
\newblock In \emph{Proceedings of the IEEE/CVF Conference on Computer Vision
  and Pattern Recognition}, pages 9737--9746, 2022.

\bibitem[Dhariwal and Nichol(2021)]{dhariwal2021diffusion}
Prafulla Dhariwal and Alexander Nichol.
\newblock Diffusion models beat gans on image synthesis.
\newblock \emph{Advances in neural information processing systems},
  34:\penalty0 8780--8794, 2021.

\bibitem[Ding et~al.(2022)Ding, Zheng, Hong, and Tang]{ding2022cogview2}
Ming Ding, Wendi Zheng, Wenyi Hong, and Jie Tang.
\newblock Cogview2: Faster and better text-to-image generation via hierarchical
  transformers.
\newblock \emph{Advances in Neural Information Processing Systems},
  35:\penalty0 16890--16902, 2022.

\bibitem[Gan et~al.(2020)Gan, Chen, Li, Zhu, Cheng, and Liu]{gan2020large}
Zhe Gan, Yen-Chun Chen, Linjie Li, Chen Zhu, Yu Cheng, and Jingjing Liu.
\newblock Large-scale adversarial training for vision-and-language
  representation learning.
\newblock \emph{Advances in Neural Information Processing Systems},
  33:\penalty0 6616--6628, 2020.

\bibitem[Gu et~al.(2022)Gu, Chen, Bao, Wen, Zhang, Chen, Yuan, and
  Guo]{gu2022vector}
Shuyang Gu, Dong Chen, Jianmin Bao, Fang Wen, Bo Zhang, Dongdong Chen, Lu Yuan,
  and Baining Guo.
\newblock Vector quantized diffusion model for text-to-image synthesis.
\newblock In \emph{Proceedings of the IEEE/CVF Conference on Computer Vision
  and Pattern Recognition}, pages 10696--10706, 2022.

\bibitem[He et~al.(2016)He, Zhang, Ren, and Sun]{he2016deep}
Kaiming He, Xiangyu Zhang, Shaoqing Ren, and Jian Sun.
\newblock Deep residual learning for image recognition.
\newblock In \emph{Proceedings of the IEEE conference on computer vision and
  pattern recognition}, pages 770--778, 2016.

\bibitem[https://www.forbes.com/sites/mattnovak/2023/03/27/ai-creates-photo-evidence-of-2001-earthquake-that-never
  happened/?sh=250435c83985(2023)]{Earthquake}
https://www.forbes.com/sites/mattnovak/2023/03/27/ai-creates-photo-evidence-of-2001-earthquake-that-never
  happened/?sh=250435c83985.
\newblock Ai creates photo evidence of 2001 earthquake that never happened.
\newblock 2023.

\bibitem[Karras et~al.(2017)Karras, Aila, Laine, and
  Lehtinen]{karras2017progressive}
Tero Karras, Timo Aila, Samuli Laine, and Jaakko Lehtinen.
\newblock Progressive growing of gans for improved quality, stability, and
  variation.
\newblock \emph{arXiv preprint arXiv:1710.10196}, 2017.

\bibitem[Karras et~al.(2019)Karras, Laine, and Aila]{karras2019style}
Tero Karras, Samuli Laine, and Timo Aila.
\newblock A style-based generator architecture for generative adversarial
  networks.
\newblock In \emph{Proceedings of the IEEE/CVF conference on computer vision
  and pattern recognition}, pages 4401--4410, 2019.

\bibitem[Li et~al.(2019)Li, Zhang, and Malik]{li2019diverse}
Ke Li, Tianhao Zhang, and Jitendra Malik.
\newblock Diverse image synthesis from semantic layouts via conditional imle.
\newblock In \emph{Proceedings of the IEEE/CVF International Conference on
  Computer Vision}, pages 4220--4229, 2019.

\bibitem[Li et~al.(2022)Li, Cai, Li, Lam, Hu, and Kot]{li2022one}
Zhi Li, Rizhao Cai, Haoliang Li, Kwok-Yan Lam, Yongjian Hu, and Alex~C Kot.
\newblock One-class knowledge distillation for face presentation attack
  detection.
\newblock \emph{IEEE Transactions on Information Forensics and Security},
  17:\penalty0 2137--2150, 2022.

\bibitem[Liu et~al.(2020)Liu, Qi, and Torr]{liu2020global}
Zhengzhe Liu, Xiaojuan Qi, and Philip~HS Torr.
\newblock Global texture enhancement for fake face detection in the wild.
\newblock In \emph{Proceedings of the IEEE/CVF conference on computer vision
  and pattern recognition}, pages 8060--8069, 2020.

\bibitem[Liu et~al.(2021)Liu, Lin, Cao, Hu, Wei, Zhang, Lin, and
  Guo]{liu2021swin}
Ze Liu, Yutong Lin, Yue Cao, Han Hu, Yixuan Wei, Zheng Zhang, Stephen Lin, and
  Baining Guo.
\newblock Swin transformer: Hierarchical vision transformer using shifted
  windows.
\newblock In \emph{Proceedings of the IEEE/CVF international conference on
  computer vision}, pages 10012--10022, 2021.

\bibitem[Nataraj et~al.(2019)Nataraj, Mohammed, Chandrasekaran, Flenner, Bappy,
  Roy-Chowdhury, and Manjunath]{nataraj2019detecting}
Lakshmanan Nataraj, Tajuddin~Manhar Mohammed, Shivkumar Chandrasekaran, Arjuna
  Flenner, Jawadul~H Bappy, Amit~K Roy-Chowdhury, and BS Manjunath.
\newblock Detecting gan generated fake images using co-occurrence matrices.
\newblock \emph{arXiv preprint arXiv:1903.06836}, 2019.

\bibitem[Nichol et~al.(2021)Nichol, Dhariwal, Ramesh, Shyam, Mishkin, McGrew,
  Sutskever, and Chen]{nichol2021glide}
Alex Nichol, Prafulla Dhariwal, Aditya Ramesh, Pranav Shyam, Pamela Mishkin,
  Bob McGrew, Ilya Sutskever, and Mark Chen.
\newblock Glide: Towards photorealistic image generation and editing with
  text-guided diffusion models.
\newblock \emph{arXiv preprint arXiv:2112.10741}, 2021.

\bibitem[Ojha et~al.(2023)Ojha, Li, and Lee]{ojha2023towards}
Utkarsh Ojha, Yuheng Li, and Yong~Jae Lee.
\newblock Towards universal fake image detectors that generalize across
  generative models.
\newblock In \emph{Proceedings of the IEEE/CVF Conference on Computer Vision
  and Pattern Recognition}, pages 24480--24489, 2023.

\bibitem[Park et~al.(2019)Park, Liu, Wang, and Zhu]{park2019semantic}
Taesung Park, Ming-Yu Liu, Ting-Chun Wang, and Jun-Yan Zhu.
\newblock Semantic image synthesis with spatially-adaptive normalization.
\newblock In \emph{Proceedings of the IEEE/CVF conference on computer vision
  and pattern recognition}, pages 2337--2346, 2019.

\bibitem[Qian et~al.(2020)Qian, Yin, Sheng, Chen, and Shao]{qian2020thinking}
Yuyang Qian, Guojun Yin, Lu Sheng, Zixuan Chen, and Jing Shao.
\newblock Thinking in frequency: Face forgery detection by mining
  frequency-aware clues.
\newblock In \emph{Computer Vision--ECCV 2020: 16th European Conference,
  Glasgow, UK, August 23--28, 2020, Proceedings, Part XII}, pages 86--103.
  Springer, 2020.

\bibitem[Radford et~al.(2021)Radford, Kim, Hallacy, Ramesh, Goh, Agarwal,
  Sastry, Askell, Mishkin, Clark, et~al.]{radford2021learning}
Alec Radford, Jong~Wook Kim, Chris Hallacy, Aditya Ramesh, Gabriel Goh,
  Sandhini Agarwal, Girish Sastry, Amanda Askell, Pamela Mishkin, Jack Clark,
  et~al.
\newblock Learning transferable visual models from natural language
  supervision.
\newblock In \emph{International conference on machine learning}, pages
  8748--8763. PMLR, 2021.

\bibitem[Ramesh et~al.(2021)Ramesh, Pavlov, Goh, Gray, Voss, Radford, Chen, and
  Sutskever]{ramesh2021zero}
Aditya Ramesh, Mikhail Pavlov, Gabriel Goh, Scott Gray, Chelsea Voss, Alec
  Radford, Mark Chen, and Ilya Sutskever.
\newblock Zero-shot text-to-image generation.
\newblock In \emph{International Conference on Machine Learning}, pages
  8821--8831. PMLR, 2021.

\bibitem[Rombach et~al.(2021)Rombach, Blattmann, Lorenz, Esser, and
  Ommer]{rombach2021highresolution}
Robin Rombach, Andreas Blattmann, Dominik Lorenz, Patrick Esser, and Björn
  Ommer.
\newblock High-resolution image synthesis with latent diffusion models, 2021.

\bibitem[Rombach et~al.(2022)Rombach, Blattmann, Lorenz, Esser, and
  Ommer]{rombach2022high}
Robin Rombach, Andreas Blattmann, Dominik Lorenz, Patrick Esser, and Bj{\"o}rn
  Ommer.
\newblock High-resolution image synthesis with latent diffusion models.
\newblock In \emph{Proceedings of the IEEE/CVF conference on computer vision
  and pattern recognition}, pages 10684--10695, 2022.

\bibitem[Rossler et~al.(2019)Rossler, Cozzolino, Verdoliva, Riess, Thies, and
  Nie{\ss}ner]{rossler2019faceforensics++}
Andreas Rossler, Davide Cozzolino, Luisa Verdoliva, Christian Riess, Justus
  Thies, and Matthias Nie{\ss}ner.
\newblock Faceforensics++: Learning to detect manipulated facial images.
\newblock In \emph{Proceedings of the IEEE/CVF international conference on
  computer vision}, pages 1--11, 2019.

\bibitem[Salehi et~al.(2021)Salehi, Sadjadi, Baselizadeh, Rohban, and
  Rabiee]{salehi2021multiresolution}
Mohammadreza Salehi, Niousha Sadjadi, Soroosh Baselizadeh, Mohammad~H Rohban,
  and Hamid~R Rabiee.
\newblock Multiresolution knowledge distillation for anomaly detection.
\newblock In \emph{Proceedings of the IEEE/CVF conference on computer vision
  and pattern recognition}, pages 14902--14912, 2021.

\bibitem[Touvron et~al.(2021)Touvron, Cord, Douze, Massa, Sablayrolles, and
  J{\'e}gou]{touvron2021training}
Hugo Touvron, Matthieu Cord, Matthijs Douze, Francisco Massa, Alexandre
  Sablayrolles, and Herv{\'e} J{\'e}gou.
\newblock Training data-efficient image transformers \& distillation through
  attention.
\newblock In \emph{International conference on machine learning}, pages
  10347--10357. PMLR, 2021.

\bibitem[Tran et~al.(2021)Tran, Tran, Nguyen, Nguyen, and Cheung]{tran2021data}
Ngoc-Trung Tran, Viet-Hung Tran, Ngoc-Bao Nguyen, Trung-Kien Nguyen, and
  Ngai-Man Cheung.
\newblock On data augmentation for gan training.
\newblock \emph{IEEE Transactions on Image Processing}, 30:\penalty0
  1882--1897, 2021.

\bibitem[Verdoliva et~al.()Verdoliva, Cozzolino, and Nagano]{verdoliva2022}
Luisa Verdoliva, Davide Cozzolino, and Koki Nagano.
\newblock 2022 ieee image and video processing cup synthetic image detection.

\bibitem[Wang et~al.(2019)Wang, Gong, and Liu]{wang2019improving}
Dilin Wang, Chengyue Gong, and Qiang Liu.
\newblock Improving neural language modeling via adversarial training.
\newblock In \emph{International Conference on Machine Learning}, pages
  6555--6565. PMLR, 2019.

\bibitem[Wang et~al.(2020)Wang, Wang, Zhang, Owens, and Efros]{wang2020cnn}
Sheng-Yu Wang, Oliver Wang, Richard Zhang, Andrew Owens, and Alexei~A Efros.
\newblock Cnn-generated images are surprisingly easy to spot... for now.
\newblock In \emph{Proceedings of the IEEE/CVF conference on computer vision
  and pattern recognition}, pages 8695--8704, 2020.

\bibitem[Wang et~al.(2023)Wang, Bao, Zhou, Wang, Hu, Chen, and
  Li]{wang2023dire}
Zhendong Wang, Jianmin Bao, Wengang Zhou, Weilun Wang, Hezhen Hu, Hong Chen,
  and Houqiang Li.
\newblock Dire for diffusion-generated image detection.
\newblock \emph{arXiv preprint arXiv:2303.09295}, 2023.

\bibitem[Xi et~al.(2023)Xi, Huang, Wei, Luo, and Zheng]{xi2023ai}
Ziyi Xi, Wenmin Huang, Kangkang Wei, Weiqi Luo, and Peijia Zheng.
\newblock Ai-generated image detection using a cross-attention enhanced
  dual-stream network.
\newblock \emph{arXiv preprint arXiv:2306.07005}, 2023.

\bibitem[Xie et~al.(2020)Xie, Tan, Gong, Wang, Yuille, and
  Le]{xie2020adversarial}
Cihang Xie, Mingxing Tan, Boqing Gong, Jiang Wang, Alan~L Yuille, and Quoc~V
  Le.
\newblock Adversarial examples improve image recognition.
\newblock In \emph{Proceedings of the IEEE/CVF conference on computer vision
  and pattern recognition}, pages 819--828, 2020.

\bibitem[Zhang et~al.(2019)Zhang, Karaman, and Chang]{zhang2019detecting}
Xu Zhang, Svebor Karaman, and Shih-Fu Chang.
\newblock Detecting and simulating artifacts in gan fake images.
\newblock In \emph{2019 IEEE international workshop on information forensics
  and security (WIFS)}, pages 1--6. IEEE, 2019.

\bibitem[Zhu et~al.(2017)Zhu, Park, Isola, and Efros]{zhu2017unpaired}
Jun-Yan Zhu, Taesung Park, Phillip Isola, and Alexei~A Efros.
\newblock Unpaired image-to-image translation using cycle-consistent
  adversarial networks.
\newblock In \emph{Proceedings of the IEEE international conference on computer
  vision}, pages 2223--2232, 2017.

\bibitem[Zhu et~al.(2023)Zhu, Chen, Yan, Huang, Lin, Li, Tu, Hu, Hu, and
  Wang]{zhu2023genimage}
Mingjian Zhu, Hanting Chen, Qiangyu Yan, Xudong Huang, Guanyu Lin, Wei Li,
  Zhijun Tu, Hailin Hu, Jie Hu, and Yunhe Wang.
\newblock Genimage: A million-scale benchmark for detecting ai-generated image.
\newblock \emph{arXiv preprint arXiv:2306.08571}, 2023.

\end{thebibliography}
}


\end{document}